\documentclass[10pt]{article} %
\usepackage[accepted]{rlc}

\usepackage{amssymb}            %
\usepackage{mathtools}          %
\usepackage{mathrsfs}           %
\usepackage{graphicx}           %
\usepackage{subcaption}         %
\usepackage[space]{grffile}     %
\usepackage{url}                %

\usepackage{booktabs} %
\usepackage{diagbox}
\usepackage{amssymb}
\usepackage{bbding} %
\usepackage{amsmath}
\usepackage{amsthm}
\usepackage{enumitem}
\usepackage[capitalize,noabbrev]{cleveref} %
\usepackage{enumitem}
\setitemize{noitemsep,topsep=0pt,parsep=0pt,partopsep=0pt}
\usepackage{algorithm}
\usepackage{algpseudocode}

\makeatletter
\newtheorem*{rep@theorem}{\rep@title}
\newcommand{\newreptheorem}[2]{%
\newenvironment{rep#1}[1]{%
 \def\rep@title{#2 \cref{##1}}%
 \begin{rep@theorem}}%
 {\end{rep@theorem}}}
\makeatother

\theoremstyle{plain}

\newreptheorem{theorem}{Theorem} %

\newreptheorem{lemma}{Lemma} %

\newreptheorem{corollary}{Corollary} %
\theoremstyle{definition}

\theoremstyle{remark}

\newcommand{\figwidthone}{0.9\textwidth}

\newcommand{\figwidththree}{0.32\textwidth}
\newcommand{\figwidthfour}{0.23\textwidth}

\newcommand{\Actions}{\mathcal{A}}
\newcommand{\States}{\mathcal{S}}
\newcommand{\Pfcn}{\mathrm{P}}

\newcommand{\Rfcn}{r}

\newcommand{\RR}{\mathbb{R}}

\newcommand{\sds}{small\_dense\_short\,}
\newcommand{\sdl}{small\_dense\_long\,}
\newcommand{\bss}{big\_sparse\_short\,}
\newcommand{\bsl}{big\_sparse\_long\,}
\newcommand{\bds}{big\_dense\_short\,}
\newcommand{\bdl}{big\_dense\_long\,}

\title{Learning to Optimize for Reinforcement Learning}

\author{Qingfeng Lan~\thanks{Work was done during an internship at Sea AI Lab, Singapore.}\\
  Department of Computing Science\\
  University of Alberta\\
  Edmonton, Alberta, Canada \\
  \texttt{qlan3@ualberta.ca} \\
  \And
  A. Rupam Mahmood \\
  Department of Computing Science \\
  University of Alberta \\
  CIFAR AI Chair, Amii \\
  \texttt{armahmood@ualberta.ca} \\
  \AND
  Shuicheng Yan \\
  Sea AI Lab, Skywork AI \\
  \texttt{shuicheng.yan@gmail.com}
  \And
  Zhongwen Xu \\
  Sea AI Lab, Tencent AI Lab \\
  \texttt{zhongwen.s.xu@gmail.com}
}

\begin{document}
\maketitle

\begin{abstract}
In recent years, by leveraging more data, computation, and diverse tasks, learned optimizers have achieved remarkable success in supervised learning, outperforming classical hand-designed optimizers.
Reinforcement learning (RL) is essentially different from supervised learning, and in practice, these learned optimizers do not work well even in simple RL tasks.
We investigate this phenomenon and identify two issues.
First, the agent-gradient distribution is non-independent and identically distributed, leading to inefficient meta-training.
Moreover, due to highly stochastic agent-environment interactions, the agent-gradients have high bias and variance, which increases the difficulty of learning an optimizer for RL.
We propose pipeline training and a novel optimizer structure with a good inductive bias to address these issues, making it possible to learn an optimizer for reinforcement learning from scratch.
We show that, although only trained in toy tasks, our learned optimizer can generalize to unseen complex tasks in Brax.~\footnote{The code is available at \url{https://github.com/sail-sg/optim4rl/}.}
\end{abstract}

\section{Introduction}

Deep learning has achieved great success in many areas~\citep{lecun2015deep}, which is largely attributed to the automatically learned features that surpass handcrafted expert features.
The use of gradient descent enables automatic adjustments of parameters within a model, yielding highly effective features. 
Despite these advancements, as another important component in deep learning, optimizers are still largely hand-designed and heavily reliant on expert knowledge.
To reduce the burden of hand-designing optimizers, researchers propose to learn to optimize with the help of meta-learning~\citep{sutton1992adapting,andrychowicz2016learning,chen2017learning,wichrowska2017learned,maheswaranathan2021reverse}.
Compared to designing optimizers with human expert knowledge, learning an optimizer is a data-driven approach, reducing the reliance on expert knowledge. 
During training, a learned optimizer can be optimized to speed learning and help achieve better performance.

Despite the significant progress in learning optimizers, previous works only present learned optimizers for supervised learning (SL).
These learned optimizers usually have complex neural network structures and incorporate numerous human-designed input features, requiring a large amount of computation and human effort to design and train them.
Moreover, they have been shown to perform poorly in reinforcement learning (RL) tasks~\citep{metz2020using,metz2022velo}.
\textit{Learning to optimize for RL remains an open and challenging problem.}

Classical optimizers are typically designed for optimization in SL tasks and then applied to RL tasks.
However, RL tasks possess unique properties that are largely overlooked by classical optimizers.
For example, unlike SL, the input distribution of an RL agent is non-stationary and non-independent and identically distributed (non-iid) due to locally correlated transition dynamics~\citep{alt2019correlation}.
Additionally, due to policy and value iterations, the target function and the loss landscapes in RL are constantly changing throughout the learning process, resulting in a much more unstable and complex optimization process.
In some cases, these properties also make it inappropriate to apply optimization algorithms designed for SL to RL directly, such as stale accumulated gradients~\citep{bengio2020correcting} or unique interference-generalization phenomenon~\citep{bengio2020interference}.
\textit{We still lack optimizers specifically designed for RL tasks.}

In this work, we aim to learn optimizers for RL.
Instead of manually designing optimizers by studying RL optimization, we apply meta-learning to learn optimizers from data generated in the agent-environment interactions.
We first investigate the problem and find that the complicated agent-gradient distribution impedes the training of learned optimizers for RL.
Furthermore, the non-iid nature of the agent-gradient distribution also hinders meta-training.
Lastly, the highly stochastic agent-environment interactions can lead to agent-gradients with high bias and variance, exacerbating the difficulty of learning an optimizer for RL.
In response to these challenges, we propose a novel approach, \textit{Optim4RL}, a learned optimizer for RL that involves pipeline training and a specialized optimizer structure of good inductive bias. 
Compared with previous methods, Optim4RL is more stable to train and more effective in optimizing RL tasks, without complex optimizer structures or 
numerous human-designed input features.
We demonstrate that Optim4RL can learn to optimize RL tasks from scratch and generalize to unseen tasks.
Our work is the first to propose a learned optimizer for deep RL tasks that works well in practice.

\section{Background}

\subsection{Reinforcement Learning}

The process of reinforcement learning (RL) can be formalized as a Markov decision process (MDP). Formally, let $M=(\States, \Actions, \Pfcn, \Rfcn, \gamma)$ be an MDP which includes a state space $\States$, an action space $\Actions$, a state transition probability function $\Pfcn: \States \times \Actions \times \States \rightarrow \RR$, a reward function $\Rfcn: \States \times \Actions \rightarrow \RR$, and a discount factor $\gamma \in [0,1)$.
At each time-step $t$, the agent observes a state $S_{t} \in \States$ and samples an action $A_{t}$ from the policy $\pi(\cdot | S_t)$.
Then it observes the next state $S_{t+1} \in \States$ according to $\Pfcn$ and receives a scalar reward $R_{t+1} = \Rfcn(S_t, A_t)$.
The return is defined as the weighted sum of rewards, i.e., $G_{t}=\sum_{k=t}^{\infty} \gamma^{k-t} R_{k+1}$.
The state-value function $v_{\pi}(s)$ is defined as the expected return starting from a state $s$.
The agent aims to find an optimal policy $\pi^*$ to maximize the expected return.

Proximal policy optimization (PPO)~\citep{schulman2017proximal} and advantage actor-critic (A2C)~\citep{mnih2016asynchronous} are two widely used RL algorithms for continuous control.
PPO improves training stability by using a clipped surrogate objective to prevent the policy from changing too much at each time step.
A2C is a variant of actor-critic method~\citep{sutton2011reinforcement}, featured with multiple-actor parallel training and synchronized gradient update.
In both algorithms, $v_{\pi}$ is approximated by $v$, which is usually parameterized as a neural network.
In practice, temporal difference (TD) learning is applied to approximate $v_{\pi}$:
\begin{equation}\label{eq:td}
v(S_t) \leftarrow v(S_t) + \alpha(R_{t+1} + \gamma v(S_{t+1}) - v(S_t)),
\end{equation}
where $\alpha$ is the learning rate, $S_t$ and $S_{t+1}$ are two successive states, and $R_{t+1} + \gamma v(S_{t+1})$ is named the TD target.
TD targets are usually biased, non-stationary, and noisy due to changing state-values, complex state transitions, and noisy reward signals~\citep{schulman2016high}.
They usually induce a changing loss landscape that evolves during training.
As a result, the agent-gradients~\footnote{In this work, we use the term agent-gradients to refer to gradients of all parameters in a learning agent, which may include policy gradient, gradient of value functions, gradient of other hyper-parameters.} usually have high bias and variance~\citep{lan2022model} which can lead to sub-optimal performance or even a failure of convergence.

\subsection{Learning to Optimize with Meta-Learning}

We aim to learn an optimizer using meta-learning.
Let $\theta$ be the agent-parameters of an RL agent that we aim to optimize.
A (learned) optimizer is defined as an update function $U$ that maps input gradients to parameter updates, implemented as a meta-network, parameterized by the meta-parameters $\phi$.
Let $z$ be the input of this meta-network which may include gradients $g$, losses $L$, exponential moving average of gradients, etc. Let $h$ be an optimizer state which stores historical values.
We can then compute agent-parameters updates $\Delta \theta$ and the updated agent-parameters $\theta'$:
\begin{equation*}
\Delta \theta, h' = U_{\phi}(z, h) \text{ and } \theta' = \theta + \Delta \theta.
\end{equation*}
Note that all classical first-order optimizers can be written in this form with $\phi=\emptyset$.
As an illustration, for SGD, $h' = h =\emptyset$, $z=g$, and $U_{\operatorname{SGD}}(g, \emptyset) = (-\alpha g, \emptyset)$, where $\alpha$ is the learning rate.
For RMSProp~\citep{tieleman2012rmsprop}, set $z=g$; $h$ is used to store the average of squared gradients. Then $U_{\operatorname{RMSProp}}(g, h) = (- \frac{\alpha g}{\sqrt{h' + \epsilon}}, h')$, where $h' = \beta h + (1-\beta) g^2$, $\beta \in [0, 1]$, and $\epsilon$ is a tiny positive number for numerical stability. 

Similar to~\citet{xu2020meta}, we apply bilevel optimization to optimize $\theta$ and $\phi$.
First, we collect $M+1$ trajectories $\mathcal{T} = \{\tau_{i}, \tau_{i+1}, \cdots, \tau_{i+M-1}, \tau_{i+M}\}$.
For the inner update, we fix $\phi$ and apply multiple steps of gradient descent updates to $\theta$ by minimizing an inner loss $L^\text{inner}$.
Specifically, for each trajectory $\tau_i \in \mathcal{T}$, we have
\begin{equation*}
\Delta \theta_i \propto \nabla_{\theta} L^\text{inner}(\tau_i; \theta_i, \phi)
\text{  and  }
\theta_{i+1} = \theta_{i} + \Delta \theta_i,
\end{equation*}
where $\nabla_{\theta} L^\text{inner}$ are agent-gradients of $\theta$.
By repeating the above process for $M$ times, we get
$\theta_i \xrightarrow{\phi} \theta_{i+1} \cdots \xrightarrow{\phi} \theta_{i+M}$.
Here, $\theta_{i+M}$ are functions of $\phi$. For simplicity, we abuse the notation and still use $\theta_{i+M}$. 
Next, we use $\tau_{i+M}$ as a validation trajectory to optimize $\phi$ with an outer loss $L^\text{outer}$:
\begin{equation*}
\Delta \phi \propto \nabla_{\phi} L^\text{outer}(\tau_{i+M}; \theta_{i+M}, \phi)
\text{  and  }
\phi' = \phi + \Delta \phi,
\end{equation*}
where $\nabla_{\phi} L^\text{outer}$ are meta-gradients of $\phi$.
Since $\theta_{i+M}$ are functions of $\phi$, we can apply the chain rule to compute meta-gradients $\nabla_{\phi} L^\text{outer}$, with the help of automatic differentiation packages.

\section{Related Work}

Our work is closely related to three areas: optimization in RL, discovering general RL algorithms, and learning to optimize in SL.

\subsection{Optimization in Reinforcement Learning}

\citet{hendersonromoff2018optimizer} tested different optimizers in RL and pointed out that classical adaptive optimizers may not always consider the complex interactions between RL algorithms and environments.
\citet{sarigul2018performance} benchmarked different momentum strategies in deep RL and found that Nesterov momentum is better at generalization.
\citet{bengio2020correcting} took one step further and showed that unlike SL, momentum in TD learning becomes doubly stale due to changing parameter updates and bootstrapping. By correcting momentum in TD learning, the sample efficiency can be improved.
\textit{These works together indicate that it may not always be appropriate to bring optimization methods in SL directly to RL without considering the unique properties in RL.}
Unlike these works which hand-design new optimizers for RL, we adopt a data-driven approach and apply meta-learning to learn an RL optimizer from data generated in the agent-environment interactions.

\subsection{Discovering General Reinforcement Learning Algorithms}

The data-driven approach is also explored in discovering general RL algorithms.
For example, \citet{houthooft2018evolved} proposed to meta-learn a loss function that takes the agent's history into account and greatly improves learning efficiency.
Similarly, \citet{kirsch2020improving} proposed MetaGenRL, which learns the objective function for deterministic policies using off-policy second-order gradients.
\citet{oh2020discovering} applied meta-learning and discovered an entire update rule for RL by interacting with a set of environments.
Instead of discovering the entire update rule, \citet{lu2022discovered} focused on exploring the mirror learning space with evolution strategies and demonstrated the generalization ability in unseen settings.
\citet{bechtle2021meta} incorporated additional information at meta-train time into parametric loss functions and applied this method to image classification, behavior cloning, and model-based RL.
\citet{kirsch2022introducing} explored the role of symmetries in discovering new RL algorithms and showed that symmetries improve generalization.
\citet{jackson2023discovering} examined the impact of environment design in meta-learning update rules in RL and developed an automatic adversarial environment design approach to improve in-distribution robustness and generalization performance of learned RL algorithms.
Following these works, our work adheres to the data-driven spirit, aiming to learn an optimizer instead of general RL algorithms.
Our training and evaluation procedures are largely inspired by them as well.

\subsection{Learning to Optimize in Supervised Learning}

Initially, learning to optimize is only applied to tune the learning rate~\citep{jacobs1988increased,sutton1992adapting,mahmood2012tuning}.
Recently, researchers started to learn an optimizer completely from scratch.
\citet{andrychowicz2016learning} implemented learned optimizers with long short-term memory networks~\citep{hochreiter1997long} and showed that learned optimizers could generalize to unseen tasks.
\citet{li2017learning} applied a guided policy search method to find a good optimizer.
\citet{wichrowska2017learned} introduced a hierarchical recurrent neural network (RNN)~\citep{medsker2001recurrent} architecture, which greatly reduces memory and computation, and was shown to generalize to different network structures.
\citet{metz2022practical} developed learned optimizers with multi-layer perceptions, which achieve a better balance among memory, computation, and performance.

Learned optimizers are known to be hard to train.
Part of the reason is that they are usually trained by truncated backpropagation through time, which leads to strongly biased gradients or exploding gradients. To overcome these issues, \citet{metz2019understanding} presented a method to dynamically weigh a reparameterization gradient estimator and an evolutionary strategy style gradient estimator, stabilizing the training of learned optimizers.
\citet{vicol2021unbiased} resolved the issues by dividing the computation graph into truncated unrolls and computing unbiased gradients with evolution strategies and gradient bias corrections.
\citet{harrison2022closer} investigated the training stability of optimization algorithms and proposed to improve the stability of learned optimizers by adding adaptive nominal terms from Adam~\citep{kingma2015adam} and AggMo~\citep{lucas2018aggregated}.
\citet{metz2020tasks} trained a general-purpose optimizer by training optimizers on thousands of tasks with a large amount of computation. 
Following the same spirit, \citet{metz2022velo} continued to perform large-scale optimizer training, leveraging more computation ($4,000$ TPU-months) and more diverse SL tasks. 
The learned optimizer, VeLO, requires no hyperparameter tuning and works well on a wide range of SL tasks. 
VeLO is the precious outcome of long-time research in the area of learning to optimize, building on the wisdom and effort of many generations.
Although marking a milestone for the success of learned optimizers in SL tasks, VeLO still performs poorly in RL tasks, as shown in Section 4.4.4 in~\citet{metz2022velo}. 

The failure of VeLO in RL tasks suggests that designing learned optimizers for RL is still a challenging problem.
Unlike previous works that focus on learning optimizers for SL, we aim to learn to optimize for RL.
As we will show next, our method is simple, stable, and effective, without using complex neural network structures or incorporating numerous human-designed features.
\textit{As far as we know, our work is the first to demonstrate the success of learned optimizers in deep RL tasks.}

\begin{figure}[tbp]
\centering
\begin{subfigure}[b]{\figwidththree}
    \centering
    \includegraphics[width=\textwidth]{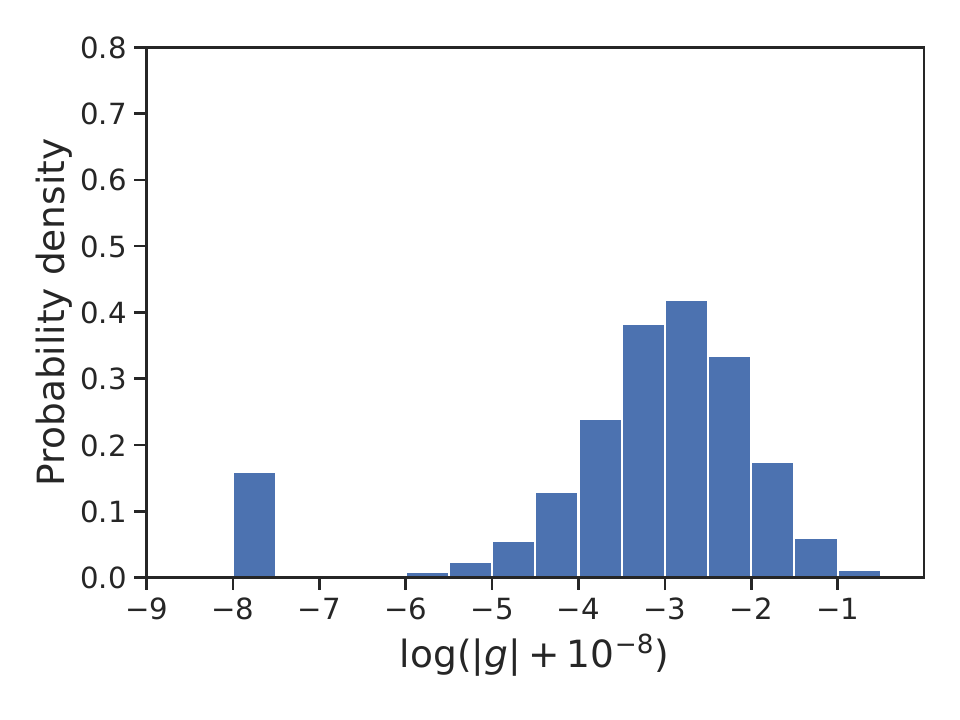}
    \caption{At the beginning of training}
\end{subfigure}
\begin{subfigure}[b]{\figwidththree}
    \centering
    \includegraphics[width=\textwidth]{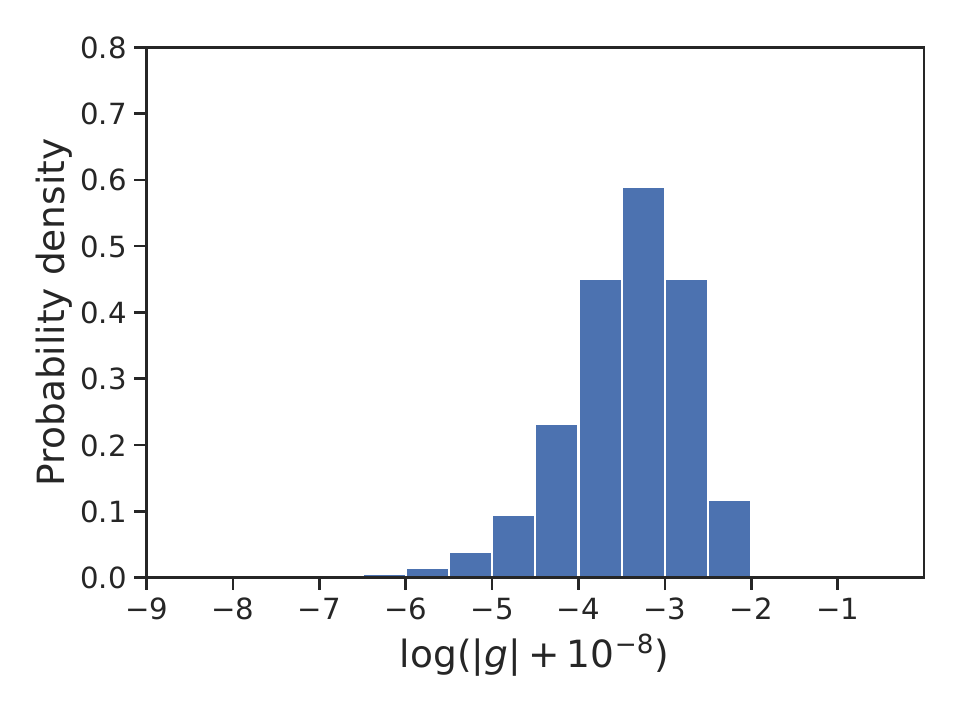}
    \caption{In the middle of training}
\end{subfigure}
\begin{subfigure}[b]{\figwidththree}
    \centering
    \includegraphics[width=\textwidth]{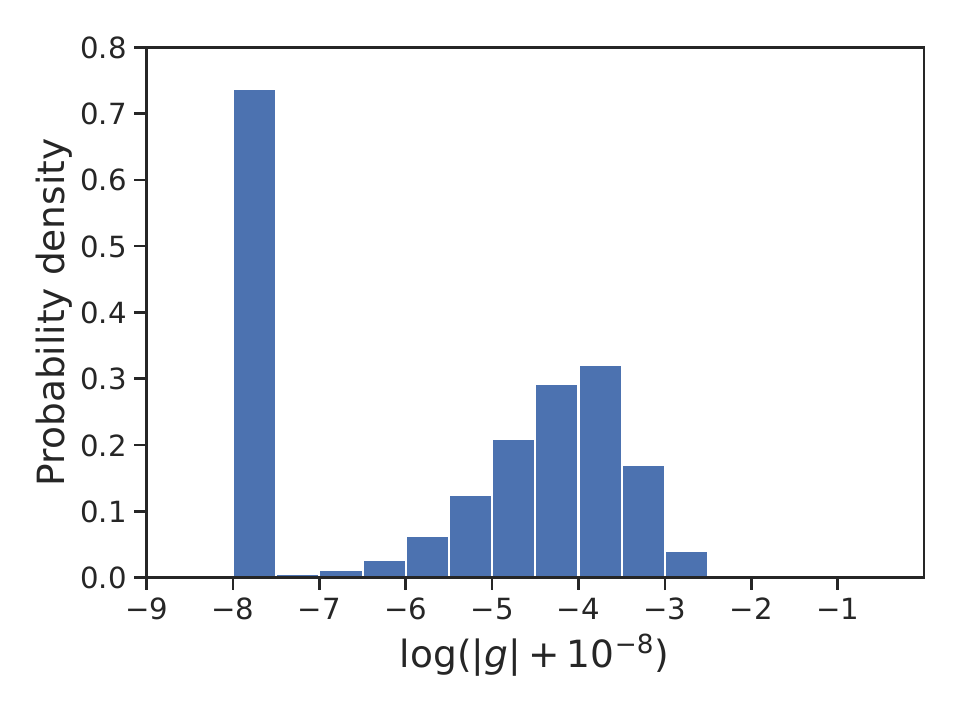}
    \caption{At the end of training}
\end{subfigure}
\caption{Visualizations of agent-gradient distributions (a) at the beginning of training, (b) in the middle of training, and (c) at the end of training. All agent-gradients are collected during training A2C in \bdl, optimized by RMSProp.
We compute $\log(|g|+10^{-8})$ to avoid the error of applying $\log$ function to non-positive agent-gradients.
}
\label{fig:grad}
\end{figure}

\section{Issues in Learning to Optimize for Reinforcement Learning}\label{sec:hardness}

Learned optimizers for SL are infamously hard to train, suffering from high training instability~\citep{wichrowska2017learned,metz2019understanding,metz2020tasks}. Learning an optimizer for RL is even harder~\citep{metz2022velo}. In the following, we identify two issues in learning to optimize for RL.

\subsection{The Agent-Gradient Distribution is Non-IID}

In RL, a learned optimizer takes the agent-gradient $g$ as an input and outputs the agent-parameter update $\Delta \theta$.
To investigate the hardness of learning an optimizer for RL, we train an A2C agent in a gridworld (i.e., \bdl, see~\cref{app:gridworld} for details) with RMSProp~\citep{tieleman2012rmsprop} and collect the agent-gradients at different training stages.
We plot these agent-gradients with logarithmic $x$-axis in~\cref{fig:grad}.
The $y$-axis shows the probability density.
Clearly, the agent-gradient distribution is non-iid, changing throughout the training process.
Specifically, at the beginning of training, there are two peaks in the agent-gradient distribution.
In the middle of training, most agent-gradients are non-zero, concentrated around $10^{-3}$.
At the end of the training, a large portion of the agent-gradients are zeros.
It is well-known that a non-iid input distribution makes training more unstable and reduces learning performance in many settings~\citep{ma2022state,wang2023comprehensive,khetarpal2022towards}.
Similarly, the violation of the iid assumption would also increase learning instability and decrease efficiency for training learned optimizers.
Note that this issue exists in both learning to optimize for SL and RL.
However, the agent-gradient distribution from RL is generally more non-iid than the gradient distribution from SL, since RL tasks are inherently more non-stationary.
For more details, please check~\cref{app:sl_gradient}.

\subsection{A Vicious Spiral of Bilevel Optimization}

Learning an optimizer while optimizing parameters of a model is a bilevel optimization, suffering from high training instability~\citep{wichrowska2017learned,metz2020tasks,harrison2022closer}.
In RL, due to highly stochastic agent-environment interactions, the agent-gradients have high bias and variance, which make the bilevel optimization even more unstable.

Specifically, in SL, it is often assumed that the training set consists of iid samples. 
However, the input data distribution in RL is non-iid, which makes the whole training process much more unstable and complex, especially when learning to optimize is involved.
In most SL settings, true labels are noiseless and time-invariant.
For example, the true label of a written digit $2$ in MNIST~\citep{deng2012mnist} is $y=2$, which does not change during training.
In RL, TD learning (see~\cref{eq:td}) is widely used, and TD targets play a similar role as labels in SL.
Unlike labels in SL, TD targets are biased, non-stationary, and noisy, due to highly stochastic agent-environment interactions.
This leads to a loss landscape that evolves during training and potentially results in the deadly triad~\citep{van2018deep} and capacity loss~\citep{lyle2021understanding}.
Moreover, in SL, a lower loss usually indicates better performance (e.g., higher classification accuracy).
But in RL, a lower outer loss is not necessarily a good indicator of better performance (i.e., higher return) due to a changing loss landscape.
Together with biased TD targets, the randomness from state transitions, reward signals, and agent-environment interactions, make the bias and variance of agent-gradients relatively high.
In learning to optimize for RL, meta-gradients are afflicted with large noise induced by the high bias and variance of agent-gradients.
With noisy and inaccurate meta-gradients, the improvement of the learned optimizer is unstable and slow.
Using a poorly performed optimizer, policy improvement is no longer guaranteed.
A poorly performed agent is unlikely to collect ``high-quality'' data to boost the performance of the agent and the learned optimizer. In the end, this bilevel optimization gets stuck in a vicious spiral: a poor optimizer $\rightarrow$ a poor agent policy $\rightarrow$ collected data of low-quality $\rightarrow$ a poor optimizer $\rightarrow$ $\cdots$.

\section{Optim4RL: A Learned Optimizer for Reinforcement Learning}

To overcome the issues in~\cref{sec:hardness}, we propose a learned optimizer for RL, named \textit{Optim4RL}, which incorporates pipeline training and a novel optimizer structure.
As we will show next, Optim4RL is more robust and efficient to train than previous methods.

\subsection{Pipeline Training}

In~\cref{fig:grad}, we show that the agent-gradient distribution is non-iid during training.
Generally, a good optimizer should be well-functioned under different agent-gradient distributions in the whole training process.
To make the agent-gradient distribution more iid, we propose \textit{pipeline training}.

\begin{figure}[tbp]
\centering
\includegraphics[width=\figwidthone]{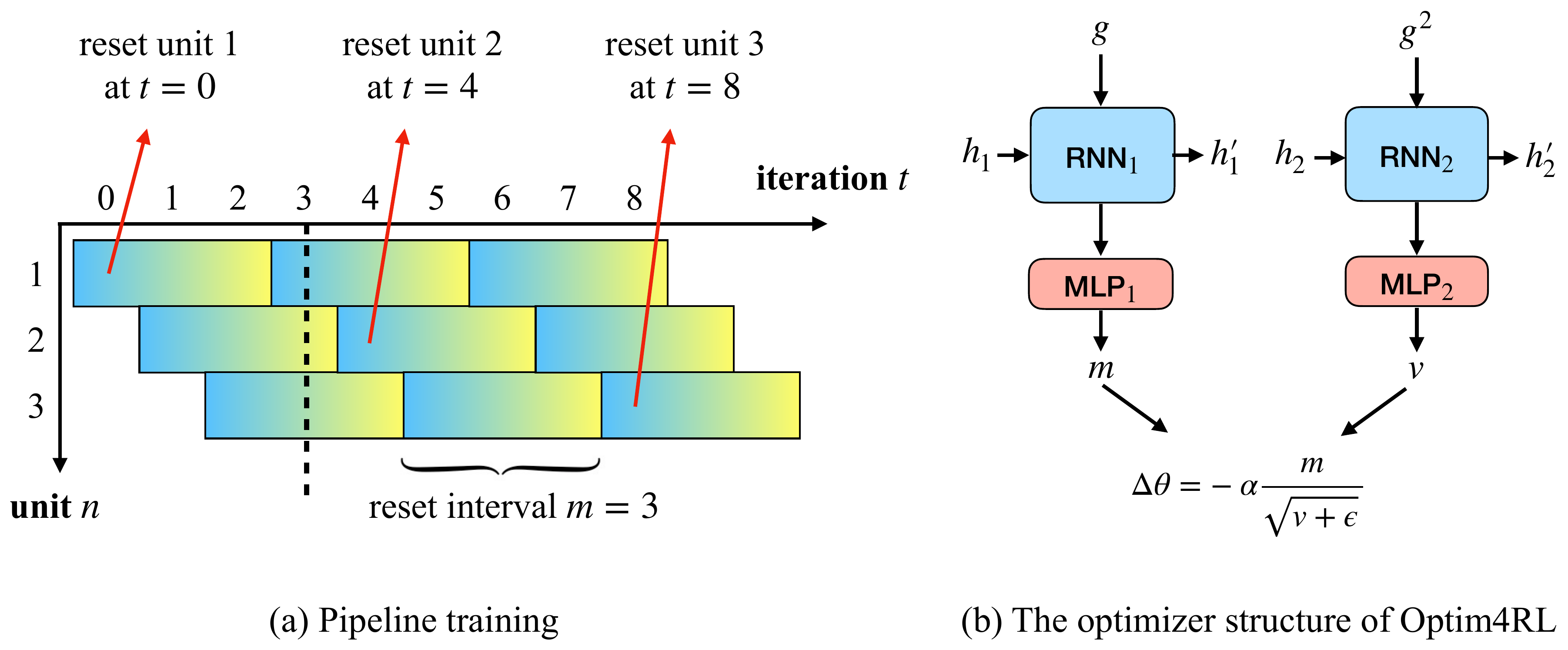}
\caption{(a) An example of pipeline training where the reset interval $m=3$ and the number of units $n=3$.
All training units are reset at regular intervals to diversify training data.
(b) The network structure of Optim4RL. $g$ is the input agent-gradient, $h_i$ and $h'_i$ are hidden states, $\alpha$ is the learning rate, $\epsilon$ is a small positive constant, and $\Delta \theta$ is the parameter update.}
\label{fig:pipeline_model}
\end{figure}

Instead of training only one agent, we train $n$ agents in parallel, each with its own task and optimizer state.
Together, the three elements form a \textit{training unit} (agent, task, optimizer state); and we have $n$ training units in total.
Let $m$ be a positive integer we call the \textit{reset interval}.
A complete \textit{training interval} lasts for $m$ training iterations.
In~\cref{fig:pipeline_model} (a), we show an example of pipeline training with $m=n=3$.
To train an optimizer effectively, the input of the learned optimizer includes agent-gradients from all $n$ training units.
Before training, we choose $n$ integers $\{r_1, \cdots, r_n\}$ such that they are evenly spaced over the interval $[0, m-1]$.
Then we assign $r_i$ to training unit $i$ for $i \in \{1, \cdots, n\}$.
At training iteration $t$, we reset training unit $i$ if $r_i \equiv t \pmod{m}$.
By resetting training units at regular intervals, it is guaranteed that at iteration $t$, we can access training data across one training interval.
For instance, at $t=3$, the input consists of agent-gradients from unit $1$ at the beginning of an interval, agent-gradients from unit $2$ at the end of an interval, and agent-gradients from unit $3$ in the middle of an interval, indicated by the dashed line in~\cref{fig:pipeline_model} (a).
With pipeline training, the input agent-gradients are more diverse and spread across a whole training interval, making the input distribution more iid.
Ideally, we expect $m \le n$ so that the input consists of agent-gradients from all training stages.
In our experiments, $n$ is the number of training environments; $m$ depends on the training steps of each task, and it has a similar magnitude as $n$.

\subsection{Improving the Inductive Bias of Learned Optimizers}\label{sec:5.2}

Recently, \citet{harrison2022closer} proved that adding adaptive terms to learned optimizers improves the training stability of optimizing a noisy quadratic model.
Experimentally, \citet{harrison2022closer} showed that adding terms from Adam~\citep{kingma2015adam} and AggMo~\citep{lucas2018aggregated} improves the stability of learned optimizers as well.
However, including human-designed features not only makes an optimizer more complex but is also against the spirit of learning to optimize --- ideal learned optimizers should be able to automatically learn useful features, reducing the reliance on human expert knowledge as much as possible.
Instead of incorporating terms from adaptive optimizers directly, we design the parameter update function in a similar form to adaptive optimizers:
\begin{equation}\label{eq:optim}
\Delta \theta = - \alpha \frac{m}{\sqrt{v+\epsilon}},
\end{equation}
where $\alpha$ is the learning rate, $\epsilon$ is a small positive number, and $m$ and $v$ are the processed outputs of dual-RNNs, as shown in~\cref{fig:pipeline_model} (b).
Specifically, for each input gradient $g$, we generate two scalars $o_1$ and $o_2$.
We then set $m = g_{sign}\exp(o_1)$ and $v = \exp(o_2)$, where $g_{sign} \in \{-1,1\}$ is the sign of $g$.
More details are included in~\cref{alg:optim4rl}.

\textit{By parameterizing the parameter update function as~\cref{eq:optim}, we improve the inductive bias of learned optimizers by choosing a suitable hypothesis space for learned optimizers and reducing the burden of approximating square root and division for neural networks.}
In general, we want to learn a good optimizer in a reasonable hypothesis space.
It should be large enough to include as many good optimizers as possible, such as Adam~\citep{kingma2015adam} and RMSProp~\citep{tieleman2012rmsprop}.
Meanwhile, it should also rule out bad choices so that a suitable candidate can be found efficiently.
An optimizer in the form of~\cref{eq:optim} meets the two requirements exactly.
Moreover, it is generally hard for neural networks to approximate mathematical operations accurately~\citep{telgarsky2017neural,yarotsky2017error,boulle2020rational,lu2021learning}.
With~\cref{eq:optim}, a neural network can spend all its expressivity and capacity learning $m$ and $v$, reducing the burden of approximating square root and division.

Finally, we combine the two techniques and propose our method --- a learned optimizer for RL (Optim4RL).
Following~\citet{andrychowicz2016learning}, our optimizer also operates coordinatewisely on agent-parameters so that all agent-parameters share the same optimizer.
Besides gradients, many previously learned optimizers for SL include human-designed features as inputs, such as moving average of gradient values at multiple timescales, moving average of squared gradients, and Adafactor-style accumulators~\citep{shazeer2018adafactor}.
In theory, these features can be learned and stored in the hidden states of RNNs in Optim4RL.
So for simplicity, we only consider agent-gradients as inputs.
As we will show next, despite its simplicity, our learned optimizer Optim4RL achieves satisfactory performance in many RL tasks, outperforming several learned optimizers.

\section{Experiment}

In this section, we first verify that Optim4RL can learn to optimize for RL from scratch.
Then, we show how to train a general-purpose learned optimizer for RL. More experimental results are included in~\cref{app:robust}.

Following~\citet{oh2020discovering}, we design several gridworlds with various properties, such as different horizons, reward functions, or state-action spaces.
More details are described in~\cref{app:gridworld}.
Besides gridworlds, we also test our method in Catch~\citep{osband2020behaviour} and Brax tasks~\citep{brax2021github}.
We mainly consider two RL algorithms --- A2C~\citep{mnih2016asynchronous} and PPO~\citep{schulman2017proximal}.
For all experiments, we train A2C in gridworlds and train PPO in Brax tasks.
For Optim4RL, due to resource constraints, we choose a small network with two GRUs~\citep{cho2014properties} of hidden size $8$; both multi-layer perceptrons (MLPs) have two hidden layers of size $16$.
We use Adam to optimize learned optimizers.
More implementation details are included in~\cref{app:detail}.

\subsection{Learning an Optimizer for RL from Scratch}\label{sec:scratch}

\begin{figure}[tbp]
\centering
\begin{subfigure}[b]{\figwidthfour}
    \centering
    \includegraphics[width=\textwidth]{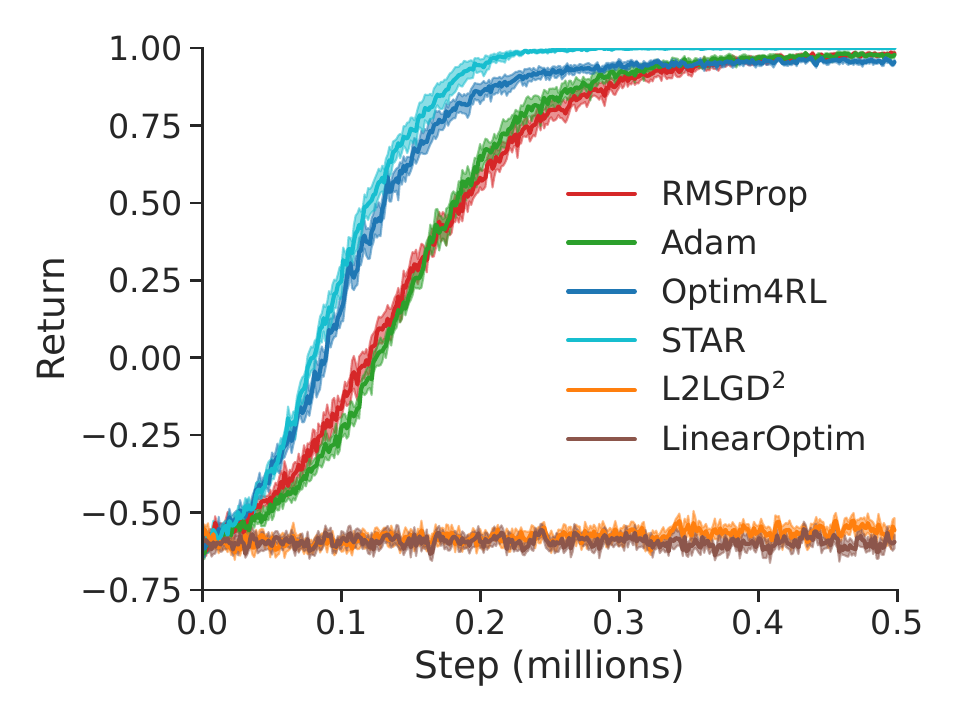}
    \caption{Catch}
\end{subfigure}
\begin{subfigure}[b]{\figwidthfour}
    \centering
    \includegraphics[width=\textwidth]{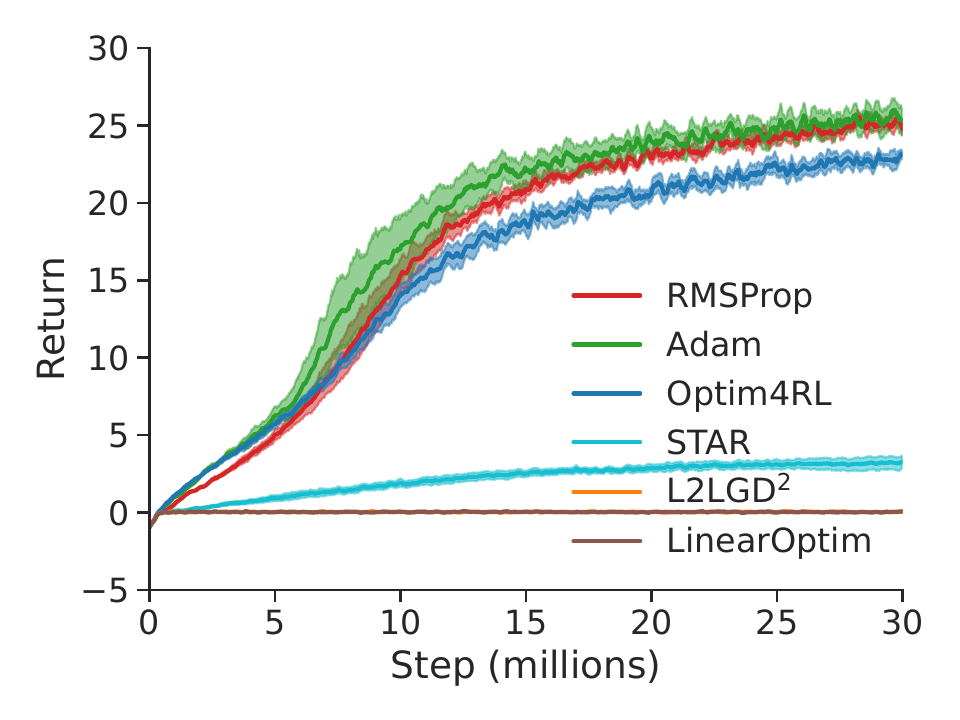}
    \caption{\bdl}
\end{subfigure}
\begin{subfigure}[b]{\figwidthfour}
    \centering
    \includegraphics[width=\textwidth]{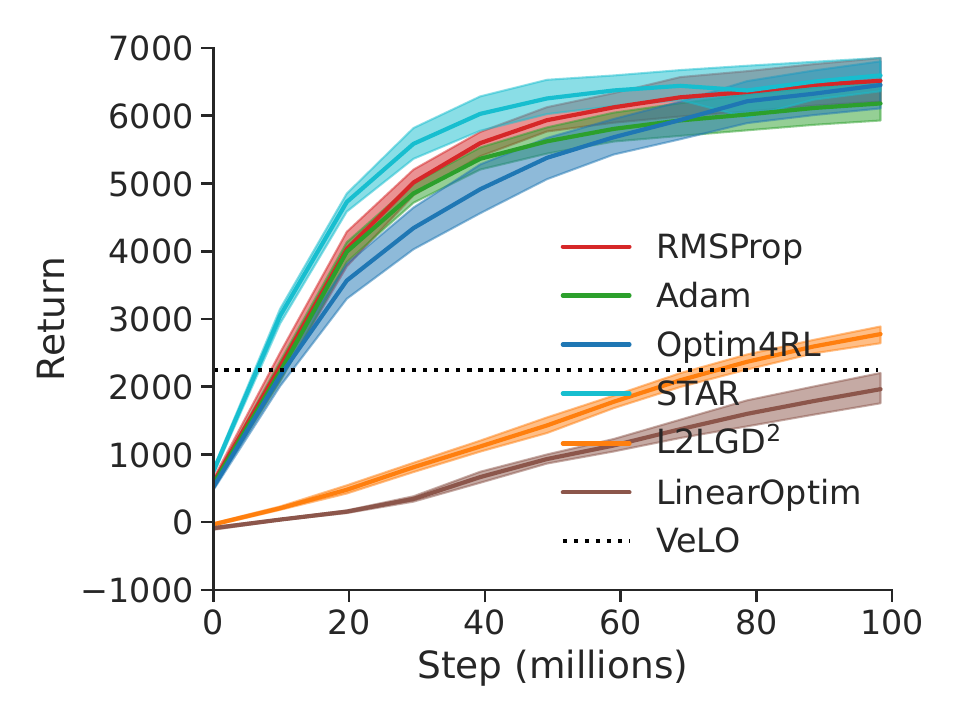}
    \caption{Ant}
\end{subfigure}
\begin{subfigure}[b]{\figwidthfour}
    \centering
    \includegraphics[width=\textwidth]{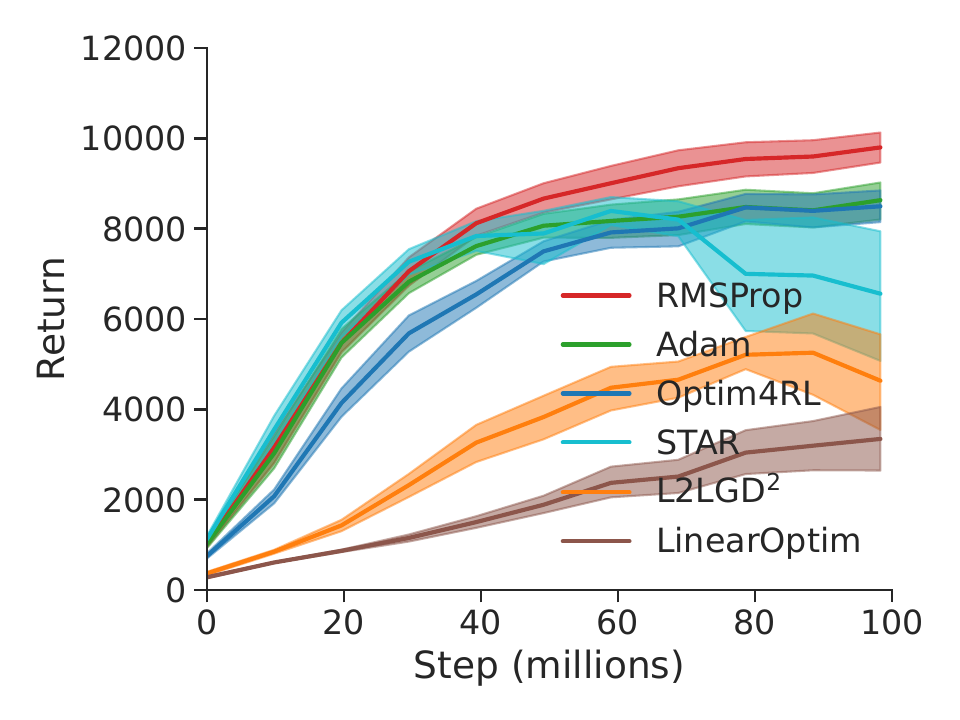}
    \caption{Humanoid}
\end{subfigure}
\caption{The optimization performance of different optimizers in four RL tasks. Note that the performance of VeLO is estimated based on Figure 11 (a) in~\citet{metz2022velo}. All other results are averaged over $10$ runs, and the shaded areas represent $90\%$ confidence intervals.
Optim4RL is the only learned optimizer that achieves satisfactory performance in all tasks.}
\label{fig:single}
\end{figure}

We first show that it is feasible to train Optim4RL in RL tasks from scratch, while learned optimizers for SL do not work well consistently in RL tasks.
We consider both classical (Adam and RMSProp) and learned optimizers ($\text{L2LGD}^2$~\citep{andrychowicz2016learning}, STAR~\citep{harrison2022closer}, and VeLO~\citep{metz2022velo}) as baselines.
Except for VeLO, we meta-learn optimizers in one task and then test the \textit{fixed} learned optimizers in this specific task.
The optimization performance of optimizers is measured by returns averaging over $10$ runs, as shown in~\cref{fig:single}.
In general, $\text{L2LGD}^2$ fails in all four tasks. In Catch, both STAR and Optim4RL perform better than classical optimizers (Adam and RMSProp), achieving a faster convergence rate.
In Ant, Optim4RL and STAR perform pretty well, on par with Adam and RMSProp, while significantly outperforming the state-of-the-art optimizer --- VeLO.
However, STAR fails to optimize effectively in \bdl; in Humanoid, STAR's performance is unstable and crashes in the end.
\textit{Optim4RL is the only learned optimizer that achieves stable and satisfactory performance in all tasks, which is a significant accomplishment in its own right, as it demonstrates the efficacy of our approach and its potential for practical applications.}

\paragraph{The advantage of the inductive bias of Optim4RL}
As an ablation study, we demonstrate the advantage of the inductive bias of OptimRL by comparing it with~\textit{LinearOptim}, which has a ``linear'' parameter update function: $\Delta \theta = -\alpha (a*g+b)$, where $\alpha$ is the learning rate, $a$ and $b$ are the outputs of an RNN model.
The only difference between LinearOptim and Optim4RL is the inductive bias --- the parameter update function of LinearOptim is in the form of a linear function. In contrast, the parameter update function of Optim4RL is inspired by adaptive optimizers (see~\cref{eq:optim}). 
As shown in~\cref{fig:single}, LinearOptim fails to optimize in all tasks, verifying the advantage of the inductive bias of Optim4RL.

\paragraph{The effectiveness of pipeline training}
By making the input agent-gradient distribution more iid and less time-dependent, pipeline training could improve the training stability and efficiency.
To verify this claim, we compare the optimization performance of Optim4RL with and without pipeline training in~\cref{tb:pipeline}.
We observe minor performance improvement in two gridworlds (\sdl and \bdl) and more significant improvement in two Brax tasks (Ant and Humanoid), confirming the effectiveness of pipeline training.

\begin{figure}[tbp]
\centering
\begin{subfigure}[b]{\figwidthfour}
    \centering
    \includegraphics[width=\textwidth]{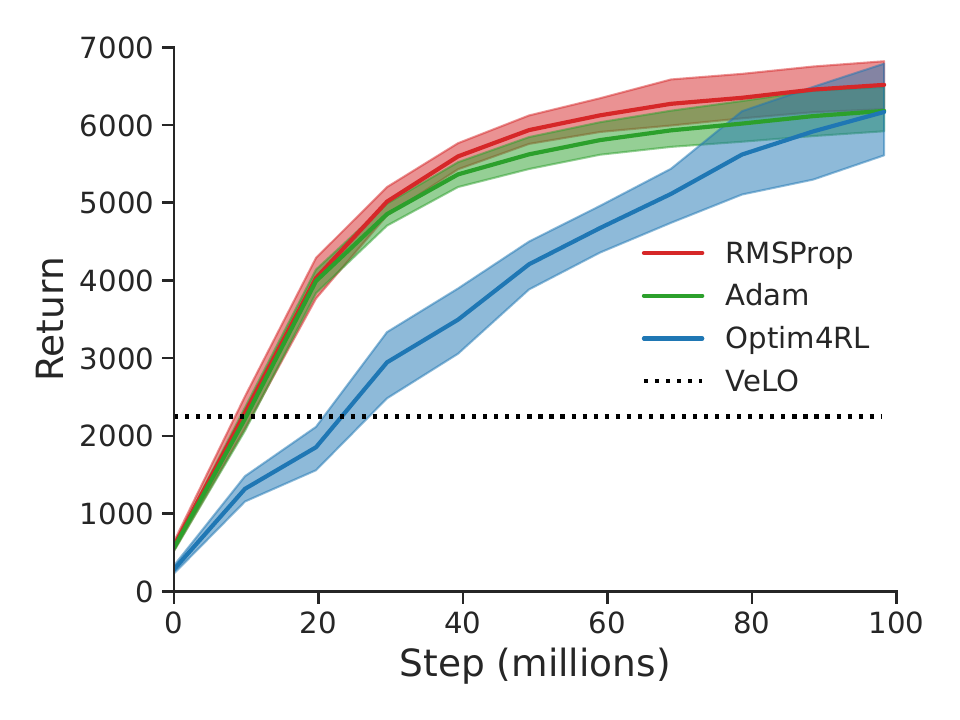}
    \caption{Ant}
\end{subfigure}
\begin{subfigure}[b]{\figwidthfour}
    \centering
    \includegraphics[width=\textwidth]{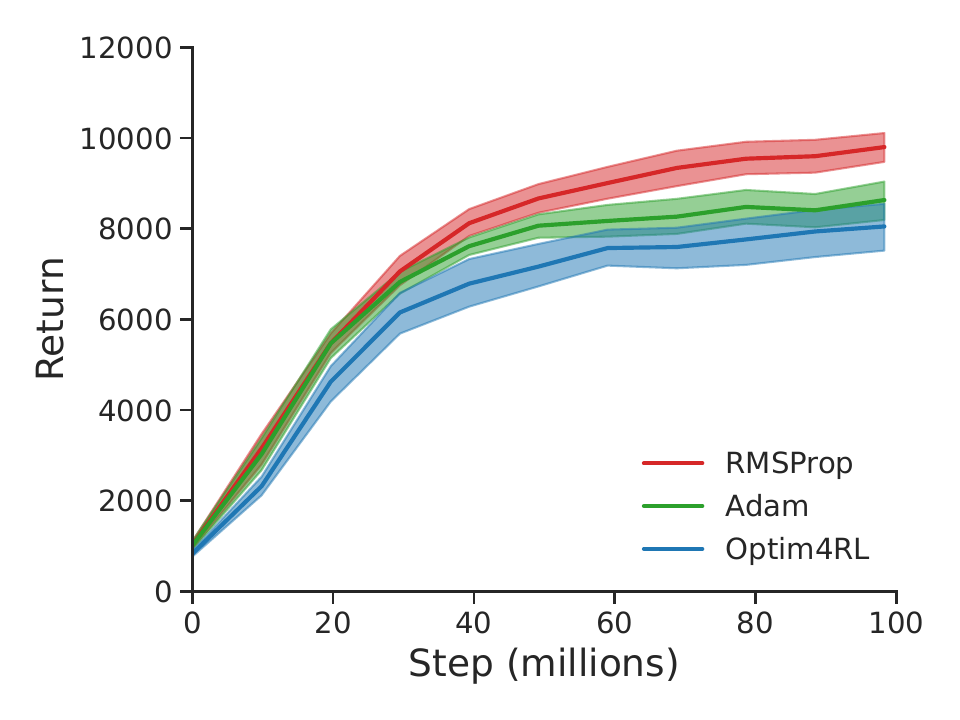}
    \caption{Humanoid}
\end{subfigure}
\begin{subfigure}[b]{\figwidthfour}
    \centering
    \includegraphics[width=\textwidth]{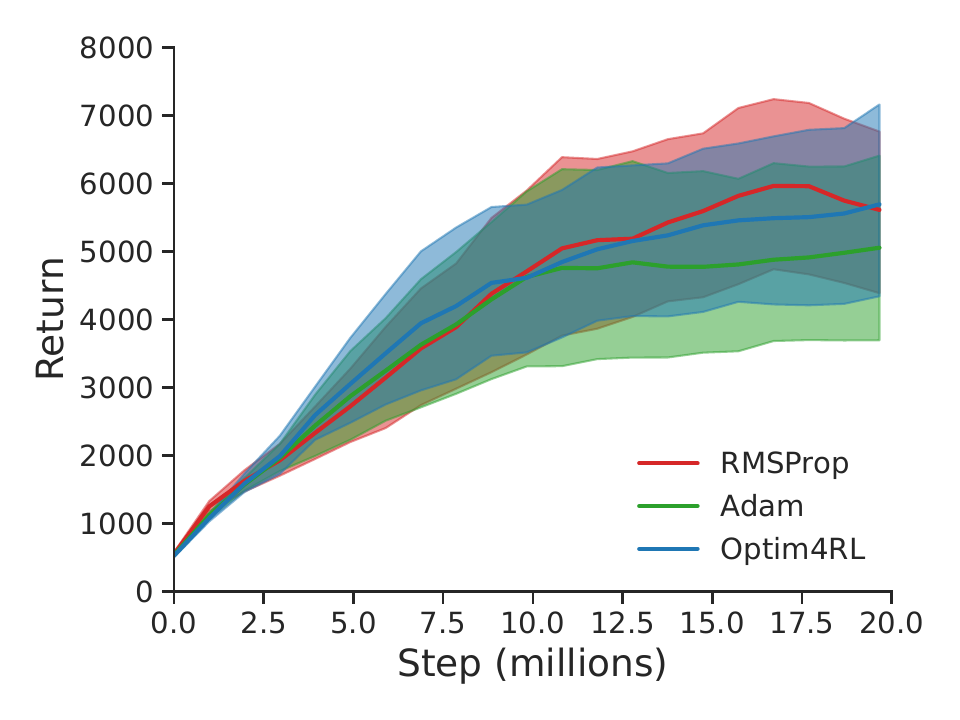}
    \caption{Pendulum}
\end{subfigure}
\begin{subfigure}[b]{\figwidthfour}
    \centering
    \includegraphics[width=\textwidth]{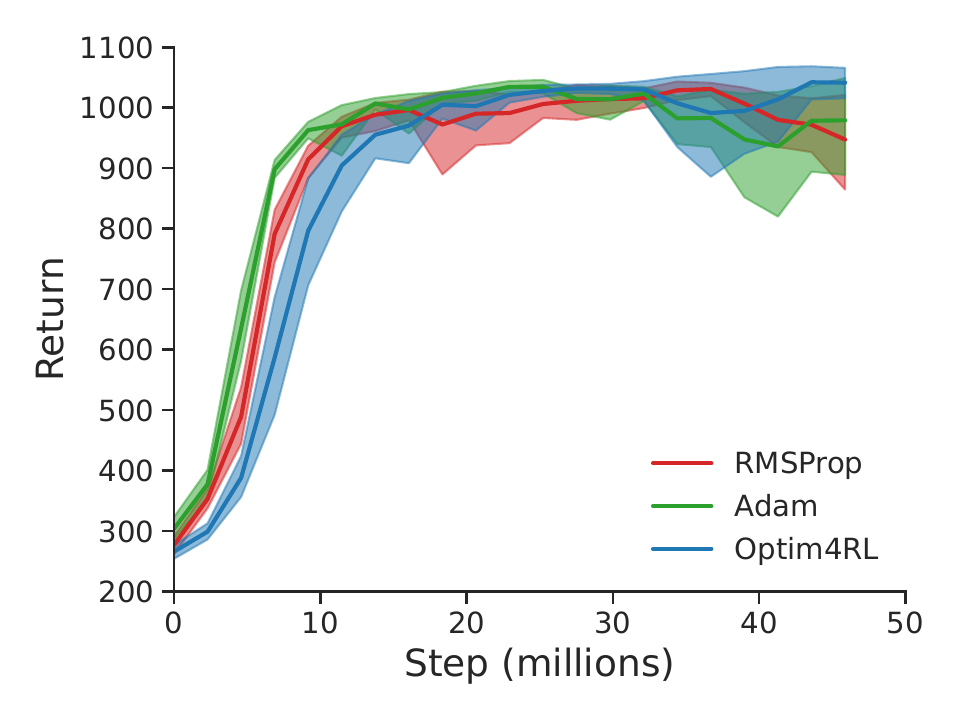}
    \caption{Walker2D}
\end{subfigure}
\caption{Optim4RL shows strong generalization ability and achieves good performance in Brax tasks, although it is only trained in six simple gridworlds from scratch.
For comparison, VeLO~\citep{metz2022velo} is trained for $4,000$ TPU-months with thousands of tasks but only achieves sub-optimal performance in Ant.
These results demonstrate the generalization ability of Optim4RL in complex unseen tasks, which is a significant achievement in itself, proving the effectiveness of our approach.
}
\label{fig:general}
\end{figure}

\begin{table}[tbp]
\centering
\begin{tabular}{lcccc}
\toprule
\diagbox{Method}{Task} & \sdl & \bdl & Ant & Humanoid \\
\midrule
With Pipeline Training & 32.22$\pm$0.52 & 23.10$\pm$0.31 & 6421$\pm$355 & 8440$\pm$364 \\
W.o. Pipeline Training & 30.64$\pm$0.69 & 22.47$\pm$0.51 & 5038$\pm$235 & 6557$\pm$1055 \\
\bottomrule
\end{tabular}
\caption{The performance of Optim4RL with and without pipeline training. All results are averaged over $10$ runs, reported with $90\%$ confidence intervals.}
\label{tb:pipeline}
\end{table}

\subsection{Toward a General-Purpose Learned Optimizer for RL}

A general-purpose optimizer should perform well even when the input gradients are at various scales.
To meta-train a learned general-purpose optimizer, first we design six gridworlds such that the generated agent-gradients in these tasks vary across a wide range.
To demonstrate the generalization ability of Optim4RL, we then meta-train Optim4RL in these gridworlds with A2C and test it in Brax tasks with PPO.
As shown in~\cref{fig:general}, Optim4RL achieves satisfactory performance in these tasks, showing a strong generalization ability.
Note that Optim4RL surpasses VeLO (the state-of-the-art learned optimizer) significantly in Ant.
This is a great success since VeLO is trained for $4,000$ TPU-months on thousands of tasks while Optim4RL is only trained in six toy tasks for a few GPU-hours.
Finally, Optim4RL is also competitive compared with classical human-designed optimizers (Adam and RMSProp), even though it is entirely trained from scratch.
\textit{Training a universally applicable learned optimizer for RL tasks is an inherently formidable challenge.
Our results demonstrate the generalization ability of Optim4RL in complex unseen tasks, which is a great achievement in itself, proving the effectiveness of our approach.}

\section{Conclusion and Future Work}

In this work, we analyzed the hardness of learning to optimize for RL and studied the failures of learned optimizers in RL. 
Our investigation reveals that agent-gradients in RL are non-iid and have high bias and variance.
To mitigate these problems, we introduced pipeline training and a novel optimizer structure.
Combining these techniques, we proposed a learned optimizer for RL, Optim4RL, which can be meta-learned to optimize RL tasks entirely from scratch.
Although only trained in toy tasks, Optim4RL showed its strong generalization ability to unseen complex tasks.

Learning to optimize for RL is a challenging problem.
Due to memory and computation constraints, our current result is limited since we can only train Optim4RL in a small number of toy tasks.
In the future, by leveraging more computation and memory, we expect to extend our approach to a larger scale and improve the performance of Optim4RL by training in more tasks with diverse RL agents.
Moreover, theoretically analyzing the convergence of learned optimizers is also an interesting topic. 
We hope our analysis and proposed method can inspire and benefit future research, paving the way for better learned optimizers for RL.

\newpage
\bibliography{reference}

\begin{thebibliography}{54}
\providecommand{\natexlab}[1]{#1}
\providecommand{\url}[1]{\texttt{#1}}
\expandafter\ifx\csname urlstyle\endcsname\relax
  \providecommand{\doi}[1]{doi: #1}\else
  \providecommand{\doi}{doi: \begingroup \urlstyle{rm}\Url}\fi

\bibitem[Alt et~al.(2019)Alt, {\v{S}}o{\v{s}}i{\'c}, and
  Koeppl]{alt2019correlation}
Bastian Alt, Adrian {\v{S}}o{\v{s}}i{\'c}, and Heinz Koeppl.
\newblock Correlation priors for reinforcement learning.
\newblock \emph{Advances in Neural Information Processing Systems}, 32, 2019.

\bibitem[Andrychowicz et~al.(2016)Andrychowicz, Denil, Gomez, Hoffman, Pfau,
  Schaul, Shillingford, and De~Freitas]{andrychowicz2016learning}
Marcin Andrychowicz, Misha Denil, Sergio Gomez, Matthew~W Hoffman, David Pfau,
  Tom Schaul, Brendan Shillingford, and Nando De~Freitas.
\newblock Learning to learn by gradient descent by gradient descent.
\newblock \emph{Advances in Neural Information Processing Systems}, 2016.

\bibitem[Bechtle et~al.(2021)Bechtle, Molchanov, Chebotar, Grefenstette,
  Righetti, Sukhatme, and Meier]{bechtle2021meta}
Sarah Bechtle, Artem Molchanov, Yevgen Chebotar, Edward Grefenstette, Ludovic
  Righetti, Gaurav Sukhatme, and Franziska Meier.
\newblock Meta learning via learned loss.
\newblock In \emph{2020 25th International Conference on Pattern Recognition
  (ICPR)}, 2021.

\bibitem[Bengio et~al.(2020{\natexlab{a}})Bengio, Pineau, and
  Precup]{bengio2020correcting}
Emmanuel Bengio, Joelle Pineau, and Doina Precup.
\newblock Correcting momentum in temporal difference learning.
\newblock \emph{NeurIPS Workshop on Deep RL}, 2020{\natexlab{a}}.

\bibitem[Bengio et~al.(2020{\natexlab{b}})Bengio, Pineau, and
  Precup]{bengio2020interference}
Emmanuel Bengio, Joelle Pineau, and Doina Precup.
\newblock Interference and generalization in temporal difference learning.
\newblock In \emph{International Conference on Machine Learning},
  2020{\natexlab{b}}.

\bibitem[Boull{\'e} et~al.(2020)Boull{\'e}, Nakatsukasa, and
  Townsend]{boulle2020rational}
Nicolas Boull{\'e}, Yuji Nakatsukasa, and Alex Townsend.
\newblock Rational neural networks.
\newblock \emph{Advances in Neural Information Processing Systems},
  33:\penalty0 14243--14253, 2020.

\bibitem[Bradbury et~al.(2018)Bradbury, Frostig, Hawkins, Johnson, Leary,
  Maclaurin, Necula, Paszke, Vander{P}las, Wanderman-{M}ilne, and
  Zhang]{jax2018github}
James Bradbury, Roy Frostig, Peter Hawkins, Matthew~James Johnson, Chris Leary,
  Dougal Maclaurin, George Necula, Adam Paszke, Jake Vander{P}las, Skye
  Wanderman-{M}ilne, and Qiao Zhang.
\newblock {JAX}: composable transformations of {P}ython+{N}um{P}y programs,
  2018.
\newblock URL \url{http://github.com/google/jax}.

\bibitem[Chen et~al.(2017)Chen, Hoffman, Colmenarejo, Denil, Lillicrap,
  Botvinick, and {de Freitas}]{chen2017learning}
Yutian Chen, Matthew~W Hoffman, Sergio~Gomez Colmenarejo, Misha Denil,
  Timothy~P Lillicrap, Matt Botvinick, and Nando {de Freitas}.
\newblock Learning to learn without gradient descent by gradient descent.
\newblock \emph{International Conference on Machine Learning}, 2017.

\bibitem[Cho et~al.(2014)Cho, van Merri{\"e}nboer, Bahdanau, and
  Bengio]{cho2014properties}
Kyunghyun Cho, Bart van Merri{\"e}nboer, Dzmitry Bahdanau, and Yoshua Bengio.
\newblock On the properties of neural machine translation: Encoder--decoder
  approaches.
\newblock In \emph{Proceedings of SSST-8, Eighth Workshop on Syntax, Semantics
  and Structure in Statistical Translation}, 2014.

\bibitem[Deng(2012)]{deng2012mnist}
Li~Deng.
\newblock The {MNIST} database of handwritten digit images for machine learning
  research.
\newblock \emph{IEEE Signal Processing Magazine}, 2012.

\bibitem[Freeman et~al.(2021)Freeman, Frey, Raichuk, Girgin, Mordatch, and
  Bachem]{brax2021github}
C.~Daniel Freeman, Erik Frey, Anton Raichuk, Sertan Girgin, Igor Mordatch, and
  Olivier Bachem.
\newblock {Brax} - a differentiable physics engine for large scale rigid body
  simulation, 2021.
\newblock URL \url{http://github.com/google/brax}.

\bibitem[Harrison et~al.(2022)Harrison, Metz, and
  Sohl-Dickstein]{harrison2022closer}
James Harrison, Luke Metz, and Jascha Sohl-Dickstein.
\newblock A closer look at learned optimization: Stability, robustness, and
  inductive biases.
\newblock In \emph{Advances in Neural Information Processing Systems}, 2022.

\bibitem[Henderson et~al.(2018)Henderson, Romoff, and
  Pineau]{hendersonromoff2018optimizer}
Peter Henderson, Joshua Romoff, and Joelle Pineau.
\newblock Where did my optimum go?: An empirical analysis of gradient descent
  optimization in policy gradient methods.
\newblock In \emph{The 14th European Workshop on Reinforcement Learning}, 2018.

\bibitem[Hochreiter \& Schmidhuber(1997)Hochreiter and
  Schmidhuber]{hochreiter1997long}
Sepp Hochreiter and J{\"u}rgen Schmidhuber.
\newblock Long short-term memory.
\newblock \emph{Neural computation}, 1997.

\bibitem[Houthooft et~al.(2018)Houthooft, Chen, Isola, Stadie, Wolski,
  Jonathan~Ho, and Abbeel]{houthooft2018evolved}
Rein Houthooft, Yuhua Chen, Phillip Isola, Bradly Stadie, Filip Wolski, OpenAI
  Jonathan~Ho, and Pieter Abbeel.
\newblock Evolved policy gradients.
\newblock \emph{Advances in Neural Information Processing Systems}, 2018.

\bibitem[Jackson et~al.(2023)Jackson, Jiang, Parker-Holder, Vuorio, Lu,
  Farquhar, Whiteson, and Foerster]{jackson2023discovering}
Matthew~Thomas Jackson, Minqi Jiang, Jack Parker-Holder, Risto Vuorio, Chris
  Lu, Gregory Farquhar, Shimon Whiteson, and Jakob~Nicolaus Foerster.
\newblock Discovering general reinforcement learning algorithms with
  adversarial environment design.
\newblock \emph{Advances in Neural Information Processing Systems}, 2023.

\bibitem[Jacobs(1988)]{jacobs1988increased}
Robert~A Jacobs.
\newblock Increased rates of convergence through learning rate adaptation.
\newblock \emph{Neural Networks}, 1988.

\bibitem[Khetarpal et~al.(2022)Khetarpal, Riemer, Rish, and
  Precup]{khetarpal2022towards}
Khimya Khetarpal, Matthew Riemer, Irina Rish, and Doina Precup.
\newblock Towards continual reinforcement learning: A review and perspectives.
\newblock \emph{Journal of Artificial Intelligence Research}, 2022.

\bibitem[Kingma \& Ba(2015)Kingma and Ba]{kingma2015adam}
Diederik~P Kingma and Jimmy Ba.
\newblock Adam: A method for stochastic optimization.
\newblock In \emph{International Conference on Learning Representations}, 2015.

\bibitem[Kirsch et~al.(2020)Kirsch, van Steenkiste, and
  Schmidhuber]{kirsch2020improving}
Louis Kirsch, Sjoerd van Steenkiste, and Juergen Schmidhuber.
\newblock Improving generalization in meta reinforcement learning using learned
  objectives.
\newblock In \emph{International Conference on Learning Representations}, 2020.

\bibitem[Kirsch et~al.(2022)Kirsch, Flennerhag, van Hasselt, Friesen, Oh, and
  Chen]{kirsch2022introducing}
Louis Kirsch, Sebastian Flennerhag, Hado van Hasselt, Abram Friesen, Junhyuk
  Oh, and Yutian Chen.
\newblock Introducing symmetries to black box meta reinforcement learning.
\newblock In \emph{Proceedings of the AAAI Conference on Artificial
  Intelligence}, 2022.

\bibitem[Lan et~al.(2022)Lan, Tosatto, Farrahi, and Mahmood]{lan2022model}
Qingfeng Lan, Samuele Tosatto, Homayoon Farrahi, and Rupam Mahmood.
\newblock Model-free policy learning with reward gradients.
\newblock In \emph{International Conference on Artificial Intelligence and
  Statistics}, 2022.

\bibitem[LeCun et~al.(2015)LeCun, Bengio, and Hinton]{lecun2015deep}
Yann LeCun, Yoshua Bengio, and Geoffrey Hinton.
\newblock Deep learning.
\newblock \emph{Nature}, 2015.

\bibitem[Li \& Malik(2017)Li and Malik]{li2017learning}
Ke~Li and Jitendra Malik.
\newblock Learning to optimize.
\newblock In \emph{International Conference on Learning Representations}, 2017.

\bibitem[Lu et~al.(2022)Lu, Kuba, Letcher, Metz, Schroeder~de Witt, and
  Foerster]{lu2022discovered}
Chris Lu, Jakub Kuba, Alistair Letcher, Luke Metz, Christian Schroeder~de Witt,
  and Jakob Foerster.
\newblock Discovered policy optimisation.
\newblock \emph{Advances in Neural Information Processing Systems}, 2022.

\bibitem[Lu et~al.(2021)Lu, Jin, Pang, Zhang, and Karniadakis]{lu2021learning}
Lu~Lu, Pengzhan Jin, Guofei Pang, Zhongqiang Zhang, and George~Em Karniadakis.
\newblock Learning nonlinear operators via {DeepONet} based on the universal
  approximation theorem of operators.
\newblock \emph{Nature Machine Intelligence}, 2021.

\bibitem[Lucas et~al.(2019)Lucas, Sun, Zemel, and Grosse]{lucas2018aggregated}
James Lucas, Shengyang Sun, Richard Zemel, and Roger Grosse.
\newblock Aggregated momentum: Stability through passive damping.
\newblock In \emph{International Conference on Learning Representations}, 2019.

\bibitem[Lyle et~al.(2021)Lyle, Rowland, and Dabney]{lyle2021understanding}
Clare Lyle, Mark Rowland, and Will Dabney.
\newblock Understanding and preventing capacity loss in reinforcement learning.
\newblock In \emph{International Conference on Learning Representations}, 2021.

\bibitem[Ma et~al.(2022)Ma, Zhu, Lin, Chen, and Qin]{ma2022state}
Xiaodong Ma, Jia Zhu, Zhihao Lin, Shanxuan Chen, and Yangjie Qin.
\newblock A state-of-the-art survey on solving non-iid data in federated
  learning.
\newblock \emph{Future Generation Computer Systems}, 2022.

\bibitem[Maheswaranathan et~al.(2021)Maheswaranathan, Sussillo, Metz, Sun, and
  Sohl-Dickstein]{maheswaranathan2021reverse}
Niru Maheswaranathan, David Sussillo, Luke Metz, Ruoxi Sun, and Jascha
  Sohl-Dickstein.
\newblock Reverse engineering learned optimizers reveals known and novel
  mechanisms.
\newblock \emph{Advances in Neural Information Processing Systems}, 2021.

\bibitem[Mahmood et~al.(2012)Mahmood, Sutton, Degris, and
  Pilarski]{mahmood2012tuning}
Ashique~Rupam Mahmood, Richard~S Sutton, Thomas Degris, and Patrick~M Pilarski.
\newblock Tuning-free step-size adaptation.
\newblock In \emph{2012 IEEE International Conference on Acoustics, Speech and
  Signal Processing (ICASSP)}, 2012.

\bibitem[Medsker \& Jain(2001)Medsker and Jain]{medsker2001recurrent}
Larry~R Medsker and LC~Jain.
\newblock Recurrent neural networks.
\newblock \emph{Design and Applications}, 2001.

\bibitem[Metz et~al.(2019)Metz, Maheswaranathan, Nixon, Freeman, and
  Sohl-Dickstein]{metz2019understanding}
Luke Metz, Niru Maheswaranathan, Jeremy Nixon, Daniel Freeman, and Jascha
  Sohl-Dickstein.
\newblock Understanding and correcting pathologies in the training of learned
  optimizers.
\newblock In \emph{International Conference on Machine Learning}, 2019.

\bibitem[Metz et~al.(2020{\natexlab{a}})Metz, Maheswaranathan, Freeman, Poole,
  and Sohl-Dickstein]{metz2020tasks}
Luke Metz, Niru Maheswaranathan, C~Daniel Freeman, Ben Poole, and Jascha
  Sohl-Dickstein.
\newblock Tasks, stability, architecture, and compute: Training more effective
  learned optimizers, and using them to train themselves.
\newblock \emph{arXiv preprint arXiv:2009.11243}, 2020{\natexlab{a}}.

\bibitem[Metz et~al.(2020{\natexlab{b}})Metz, Maheswaranathan, Sun, Freeman,
  Poole, and Sohl-Dickstein]{metz2020using}
Luke Metz, Niru Maheswaranathan, Ruoxi Sun, C~Daniel Freeman, Ben Poole, and
  Jascha Sohl-Dickstein.
\newblock Using a thousand optimization tasks to learn hyperparameter search
  strategies.
\newblock \emph{arXiv preprint arXiv:2002.11887}, 2020{\natexlab{b}}.

\bibitem[Metz et~al.(2022{\natexlab{a}})Metz, Freeman, Harrison,
  Maheswaranathan, and Sohl-Dickstein]{metz2022practical}
Luke Metz, C~Daniel Freeman, James Harrison, Niru Maheswaranathan, and Jascha
  Sohl-Dickstein.
\newblock Practical tradeoffs between memory, compute, and performance in
  learned optimizers.
\newblock In \emph{Conference on Lifelong Learning Agents}, 2022{\natexlab{a}}.

\bibitem[Metz et~al.(2022{\natexlab{b}})Metz, Harrison, Freeman, Merchant,
  Beyer, Bradbury, Agrawal, Poole, Mordatch, Roberts, et~al.]{metz2022velo}
Luke Metz, James Harrison, C~Daniel Freeman, Amil Merchant, Lucas Beyer, James
  Bradbury, Naman Agrawal, Ben Poole, Igor Mordatch, Adam Roberts, et~al.
\newblock {VeLO}: Training versatile learned optimizers by scaling up.
\newblock \emph{arXiv preprint arXiv:2211.09760}, 2022{\natexlab{b}}.

\bibitem[Mnih et~al.(2016)Mnih, Badia, Mirza, Graves, Lillicrap, Harley,
  Silver, and Kavukcuoglu]{mnih2016asynchronous}
Volodymyr Mnih, Adria~Puigdomenech Badia, Mehdi Mirza, Alex Graves, Timothy
  Lillicrap, Tim Harley, David Silver, and Koray Kavukcuoglu.
\newblock Asynchronous methods for deep reinforcement learning.
\newblock In \emph{International conference on machine learning}, 2016.

\bibitem[Oh et~al.(2020)Oh, Hessel, Czarnecki, Xu, van Hasselt, Singh, and
  Silver]{oh2020discovering}
Junhyuk Oh, Matteo Hessel, Wojciech~M Czarnecki, Zhongwen Xu, Hado~P van
  Hasselt, Satinder Singh, and David Silver.
\newblock Discovering reinforcement learning algorithms.
\newblock \emph{Advances in Neural Information Processing Systems}, 2020.

\bibitem[Osband et~al.(2020)Osband, Doron, Hessel, Aslanides, Sezener, Saraiva,
  McKinney, Lattimore, Szepesvari, Singh, et~al.]{osband2020behaviour}
Ian Osband, Yotam Doron, Matteo Hessel, John Aslanides, Eren Sezener, Andre
  Saraiva, Katrina McKinney, Tor Lattimore, Csaba Szepesvari, Satinder Singh,
  et~al.
\newblock Behaviour suite for reinforcement learning.
\newblock In \emph{International Conference on Learning Representations}, 2020.

\bibitem[Sarig{\"u}l \& Avci(2018)Sarig{\"u}l and Avci]{sarigul2018performance}
Mehmet Sarig{\"u}l and Mutlu Avci.
\newblock Performance comparison of different momentum techniques on deep
  reinforcement learning.
\newblock \emph{Journal of Information and Telecommunication}, 2018.

\bibitem[Schulman et~al.(2016)Schulman, Moritz, Levine, Jordan, and
  Abbeel]{schulman2016high}
John Schulman, Philipp Moritz, Sergey Levine, Michael Jordan, and Pieter
  Abbeel.
\newblock High-dimensional continuous control using generalized advantage
  estimation.
\newblock In \emph{International Conference on Learning Representations}, 2016.

\bibitem[Schulman et~al.(2017)Schulman, Wolski, Dhariwal, Radford, and
  Klimov]{schulman2017proximal}
John Schulman, Filip Wolski, Prafulla Dhariwal, Alec Radford, and Oleg Klimov.
\newblock Proximal policy optimization algorithms.
\newblock \emph{arXiv preprint arXiv:1707.06347}, 2017.

\bibitem[Shazeer \& Stern(2018)Shazeer and Stern]{shazeer2018adafactor}
Noam Shazeer and Mitchell Stern.
\newblock Adafactor: Adaptive learning rates with sublinear memory cost.
\newblock In \emph{International Conference on Machine Learning}, 2018.

\bibitem[Sutton(1992)]{sutton1992adapting}
Richard~S Sutton.
\newblock Adapting bias by gradient descent: An incremental version of
  delta-bar-delta.
\newblock In \emph{Proceedings of the AAAI Conference on Artificial
  Intelligence}, 1992.

\bibitem[Sutton \& Barto(2018)Sutton and Barto]{sutton2011reinforcement}
Richard~S Sutton and Andrew~G Barto.
\newblock \emph{{Reinforcement Learning: An Introduction}}.
\newblock MIT Press, second edition, 2018.

\bibitem[Telgarsky(2017)]{telgarsky2017neural}
Matus Telgarsky.
\newblock Neural networks and rational functions.
\newblock In \emph{International Conference on Machine Learning}, 2017.

\bibitem[Tieleman \& Hinton(2012)Tieleman and Hinton]{tieleman2012rmsprop}
Tijmen Tieleman and Geoffrey Hinton.
\newblock Lecture 6.5-{RMSProp}: Divide the gradient by a running average of
  its recent magnitude.
\newblock \emph{COURSERA Neural Networks Neural Networks for Machine Learning},
  2012.

\bibitem[Van~Hasselt et~al.(2018)Van~Hasselt, Doron, Strub, Hessel, Sonnerat,
  and Modayil]{van2018deep}
Hado Van~Hasselt, Yotam Doron, Florian Strub, Matteo Hessel, Nicolas Sonnerat,
  and Joseph Modayil.
\newblock Deep reinforcement learning and the deadly triad.
\newblock \emph{arXiv preprint arXiv:1812.02648}, 2018.

\bibitem[Vicol et~al.(2021)Vicol, Metz, and Sohl-Dickstein]{vicol2021unbiased}
Paul Vicol, Luke Metz, and Jascha Sohl-Dickstein.
\newblock Unbiased gradient estimation in unrolled computation graphs with
  persistent evolution strategies.
\newblock In \emph{International Conference on Machine Learning}, 2021.

\bibitem[Wang et~al.(2023)Wang, Zhang, Su, and Zhu]{wang2023comprehensive}
Liyuan Wang, Xingxing Zhang, Hang Su, and Jun Zhu.
\newblock A comprehensive survey of continual learning: Theory, method and
  application.
\newblock \emph{arXiv preprint arXiv:2302.00487}, 2023.

\bibitem[Wichrowska et~al.(2017)Wichrowska, Maheswaranathan, Hoffman,
  Colmenarejo, Denil, Freitas, and Sohl-Dickstein]{wichrowska2017learned}
Olga Wichrowska, Niru Maheswaranathan, Matthew~W Hoffman, Sergio~Gomez
  Colmenarejo, Misha Denil, Nando Freitas, and Jascha Sohl-Dickstein.
\newblock Learned optimizers that scale and generalize.
\newblock In \emph{International Conference on Machine Learning}, 2017.

\bibitem[Xu et~al.(2020)Xu, van Hasselt, Hessel, Oh, Singh, and
  Silver]{xu2020meta}
Zhongwen Xu, Hado~P van Hasselt, Matteo Hessel, Junhyuk Oh, Satinder Singh, and
  David Silver.
\newblock Meta-gradient reinforcement learning with an objective discovered
  online.
\newblock \emph{Advances in Neural Information Processing Systems}, 2020.

\bibitem[Yarotsky(2017)]{yarotsky2017error}
Dmitry Yarotsky.
\newblock Error bounds for approximations with deep {ReLU} networks.
\newblock \emph{Neural Networks}, 2017.

\end{thebibliography}
\bibliographystyle{rlc}

\newpage
\appendix
\onecolumn

\section{Pseudocode of Optim4RL}

The pseudocode of Optim4RL is presented in~\cref{alg:optim4rl}.
Specifically, in all our experiments, we use GRUs~\citep{cho2014properties} with hidden size $8$, MLPs with hidden sizes $[16,16]$.

\begin{algorithm}
\caption{Optim4RL: A Learned Optimizer for Reinforcement Learning}\label{alg:optim4rl}
\begin{algorithmic}
\Require $\operatorname{RNN}_1$ and $\operatorname{RNN}_2$, $\operatorname{MLP}_1$ and $\operatorname{MLP}_2$, hidden states $h_1$ and $h_2$, input gradient $g$, $\epsilon = 10^{-8}$, learning rate $\alpha$.
\State $g \gets \perp{g}$ \Comment{$\perp$ denotes the stop-gradient operation}
\State $h_1, x_1 \gets \operatorname{RNN}_1(h_1, g)$ and $o_1 = \operatorname{MLP}_1(x_1)$
\State $m = \operatorname{sign}(g) \exp(o_1)$ \Comment{Compute $m$: 1st pseudo moment estimate}
\State $h_2, x_2 \gets \operatorname{RNN}_2(h_2, g^2)$ and $o_2 = \operatorname{MLP}_2(x_2)$
\State $v = \exp(o_2)$ \Comment{Compute $v$: 2nd pseudo moment estimate}
\State $\Delta \theta \gets -\alpha \frac{m}{\sqrt{v+\epsilon}}$ \Comment{Compute the parameter update}
\end{algorithmic}
\end{algorithm}

\section{Gridworlds}\label{app:gridworld}

We follow~\citet{oh2020discovering} and design $6$ gridwolds.
In each gridworld, there are $N$ objects.
Each object is described as $[r, \epsilon_\text{term}, \epsilon_\text{respawn}]$.
Object locations are randomly determined at the beginning of each episode, and an object reappears at a random location after being collected, with a probability of $\epsilon_\text{respawn}$ for each time-step.
The observation consists of a tensor $\{0, 1\}^{N \times H \times W}$, where $N$ is the number of objects, and $H \times W$ is the size of the grid.
An agent has $9$ movement actions for adjacent positions, including staying in the same position.
When the agent collects an object, it receives the corresponding reward ($r \times \text{reward scale}$), and the episode terminates with a probability of $\epsilon_\text{term}$ associated with the object.
The default reward scale is 1.
In~\cref{tb:grid}, we describe the setting of each gridworld in detail.

\begin{table}[htbp]
\centering
\begin{tabular}{lcccc}
\toprule
\diagbox{Task}{Setting} & Size ($H \times W$) & Objects & Horizon \\
\midrule
\bss & $10 \times 12$ & $2 \times [1.0, 0.0, 0.05]$, $2 \times [-1.0, 0.5, 0.05]$ & $50$ \\
\bsl & $12 \times 10$ & $2 \times [1.0, 0.0, 0.05]$, $2 \times [-1.0, 0.5, 0.05]$ & $500$ \\
\bds & $9 \times 13$ & $2 \times [1.0, 0.0, 0.5]$, $2 \times [-1.0, 0.5, 0.5]$ & $50$ \\
\bdl & $13 \times 9$ & $2 \times [1.0, 0.0, 0.5]$, $2 \times [-1.0, 0.5, 0.5]$ & $500$ \\
\sdl & $6 \times 4$ & $[1.0, 0.0, 0.5]$, $[-1.0, 0.5, 0.5]$ & $500$ \\
\sds & $4 \times 6$ & $[1.0, 0.0, 0.5]$, $[-1.0, 0.5, 0.5]$ & $50$ \\
\bottomrule
\end{tabular}
\caption{The detailed settings of gridworlds.}
\label{tb:grid}
\end{table}

\section{Experimental Details}\label{app:detail}

In this work, we apply Jax~\citep{jax2018github} to do automatic differentiation.
For A2C training in gridworlds, the feature net is an MLP with hidden size $32$ for the ``small'' gridworlds.
For the ``big'' gridworlds, the feature net is a convolution neural network (CNN) with $16$ features and kernel size $2$, followed by an MLP with output size $32$.
Unless mentioned explicitly, we use $\operatorname{ReLU}$ as the activation function.
We set $\lambda=0.95$ to compute $\lambda$-returns.
The discount factor $\gamma=0.995$.
One rollout has $20$ steps.
The critic loss weight is $0.5$, and the entropy weight is $0.01$.  
For PPO training in Brax games, we use the same settings in Brax examples~\footnote{\url{https://github.com/google/brax/blob/main/notebooks/training.ipynb}}.
To meta-learn optimizers, we set $M=4$ in all experiments; that is, for every outer update, we do $4$ inner updates. Potentially, a larger $M$ could lead to more farsighted learning but results in increasing memory and computation requirements. We set $M=4$ as a trade-off, which also works well in practice.
Following common practice~\citep{lu2022discovered}, we report results averaged over $10$ runs.
Other details are presented in the following sections.

\subsection{Computation Resource}

All our experiments can be trained with V100 GPUs. For some experiments, we use $4$ V100 GPUs due to a large GPU memory requirement.
The computation to repeat all experimental results in this work should be less than $1$ GPU-year, while the exact computation used is hard to estimate.

\subsection{Implementation Details for Section 4}

We collect agent-gradients by training A2C in \bdl for $30M$ steps with learning rate $3e-3$, optimized by RMSProp.
All collected agent-gradients are divided into $30$ parts by time-steps.
We then plot the agent-gradients in the first, sixteenth, and last parts as the agent-gradient distributions at the beginning, middle, and end of training, respectively.

\subsection{Implementation Details for Section 6.1}

For both LinearOptim and $\text{L2LGD}^2$, the model consists of a GRU with hidden size $8$, followed by an MLP with hidden sizes $[16,16]$.
For STAR, we use the official implementation from learned\_optimization~\footnote{\url{https://github.com/google/learned_optimization/blob/main/learned_optimization/learned_optimizers/adafac_nominal.py}}.
Unlike the supervised learning setting, we set weight decay to $0$ since a positive weight decay in STAR leads to much worse performance.
For a fair comparison, we apply pipeline training to train all learned optimizers.

For Catch, the agent learning rate is $1e-3$; the number of environments / training units $n$ is $64$; the reset interval $m$ is chosen from $\{32, 64\}$.
For \bdl, the agent learning rate is $3e-3$; the number of environments / training units $n$ is $512$; the reset interval $m$ is chosen from $\{72, 144, 288, 576\}$.
For Ant and Humanoid, the agent learning rate is $3e-4$; the number of environments / training units $n$ is $2048$; the reset interval $m$ is chosen from $\{32, 64, 128, 256, 512\}$.
Furthermore, in order to reduce memory requirement, we set the number of mini-batches to $8$; and change the hidden sizes of the value network from $[256,256,256,256,256]$ to $[64,64,64,64,64]$.

We use Adam as the meta optimizer and choose the meta learning rate from $\{1e-5, 3e-5, 1e-4, 3e-4, 1e-3, 3e-3, 1e-2\}$.

\subsection{Implementation Details for Section 6.2}

Optim4RL is meta-trained in $6$ gridworlds and then tested in Brax tasks.
We use Adam as the meta optimizer and choose the meta learning rate from $\{1e-5, 3e-5, 1e-4, 3e-4, 1e-3, 3e-3\}$.
The number of environments/training units $n$ is $512$.
The reset interval $m$ is chosen from $\{72, 144, 288, 576\}$.
The reward scales of all gridworlds are in~\cref{tb:reward}.

\begin{table}[htbp]
\centering
\begin{tabular}{lcc}
\toprule
Gridworld & Reward Scale \\
\midrule
\sdl & 1000 \\
\sds & 100  \\
\bss & 100  \\
\bds & 10   \\
\bsl & 10   \\
\bdl & 1    \\
\bottomrule
\end{tabular}
\caption{The reward scales of gridworlds used for learning a general-purpose optimizer.}
\label{tb:reward}
\end{table}

\section{Experiments for the Gradient Distribution in Supervised Learning}\label{app:sl_gradient}

\begin{figure}[htbp]
\centering
\begin{subfigure}[b]{\figwidththree}
    \centering
    \includegraphics[width=\textwidth]{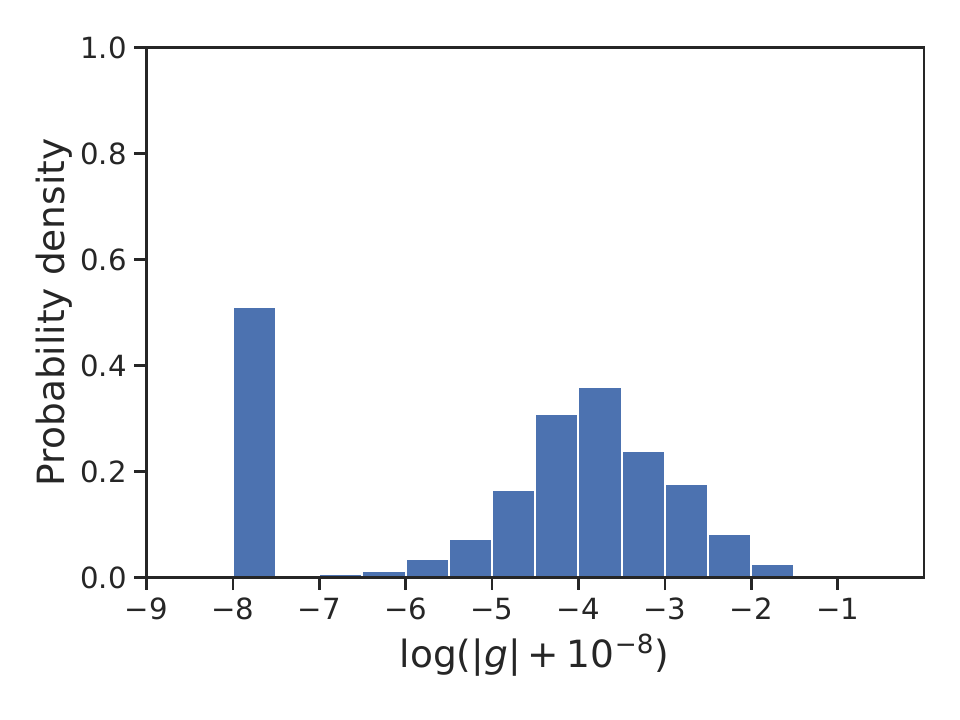}
    \caption{At the beginning of training}
\end{subfigure}
\begin{subfigure}[b]{\figwidththree}
    \centering
    \includegraphics[width=\textwidth]{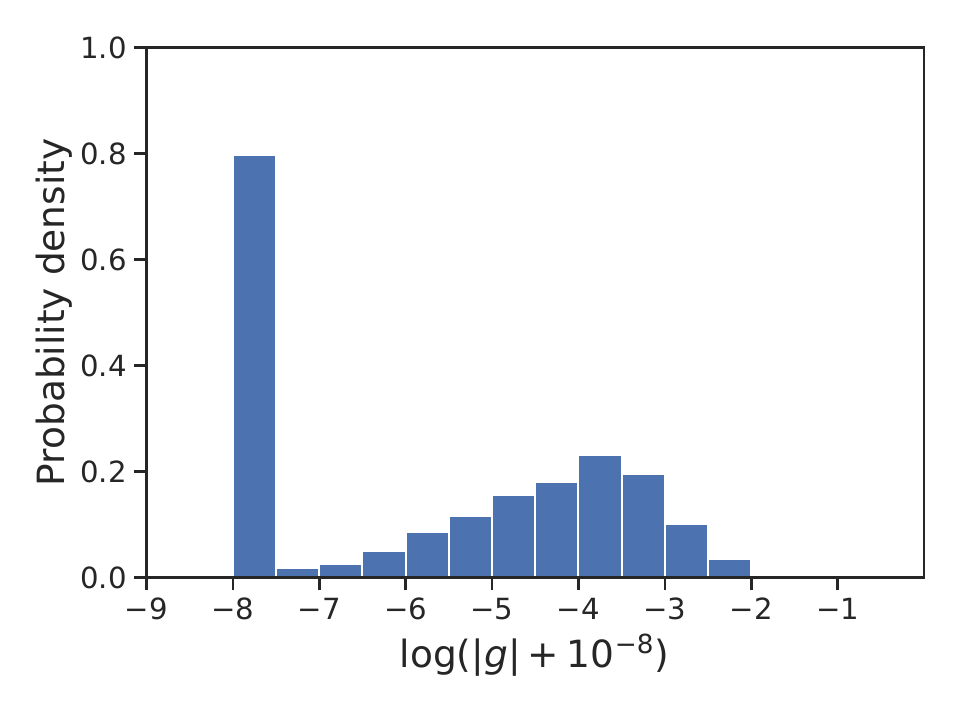}
    \caption{In the middle of training}
\end{subfigure}
\begin{subfigure}[b]{\figwidththree}
    \centering
    \includegraphics[width=\textwidth]{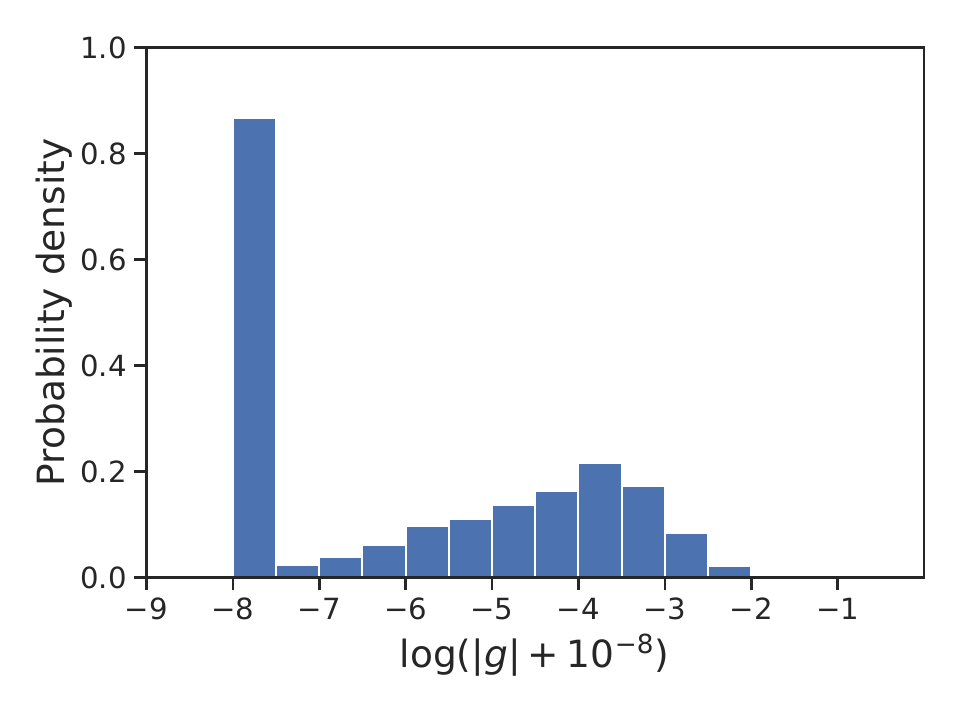}
    \caption{At the end of training}
\end{subfigure}
\caption{Visualizations of gradient distributions (a) at the beginning of training, (b) in the middle of training, and (3) at the end of training. All gradients are collected during training in MNIST, optimized by RMSProp.}
\label{fig:grad_sl}
\end{figure}

In this section, we show that the gradient distribution in supervised learning is also non-iid, but it is more iid than the agent-gradient distribution in RL. 
Specifically, we train a neural network on MNIST~\citep{deng2012mnist} for $10$ epochs with RMSProp and collect gradients at different training stages. Note that the network is the same as the actor network used in training A2C in \bdl, except for the output layer.
We plot these gradients with logarithmic $x$-axis in~\cref{fig:grad_sl}.
Similar to~\cref{fig:grad}, the gradient distribution is also non-iid, changing throughout the training process.

To show the gradient distribution from training on MNIST is more iid than the agent-gradient distribution from training in \bdl, we compute the Wasserstein distance (WD) between the (agent-)gradient distribution at different training stages and the distribution of all (agent-)gradients during training in~\cref{tb:wd}. Note that a smaller distance indicates a higher iid degree. Thus these results support the above claim. 

\begin{table}[htbp]
\centering
\begin{tabular}{lccc}
\toprule
\diagbox{Task}{WD Value} & WD(beginning, all) & WD(middle, all) & WD(end, all) \\
\midrule 
MNIST & $4.0642 \times 10^{-4}$  &  $0.7205 \times 10^{-4}$  &  $1.4204 \times 10^{-4}$ \\
\bdl  & $7.1583 \times 10^{-4}$  &  $1.1726 \times 10^{-4}$  &  $6.7967 \times 10^{-4}$ \\
\bottomrule
\end{tabular}
\caption{The Wasserstein distance between the (agent-)gradient distribution at different training stages and the distribution of all (agent-)gradients during training.}
\label{tb:wd}
\end{table}

\section{Robust Training and Strong Generalization Under Different Hyper-Parameter Settings}\label{app:robust}

Generally, we find it hard to train learned optimizers partly due to Not a Number (NaN) errors during training, even when gradient clipping or gradient normalization is applied.
For example, among all meta-training hyper-parameter settings, we fail to train STAR due to NaN errors in more than $80\%$ and $50\%$ settings in Humanoid and Ant, respectively.
However, NaN errors are seldom encountered when we meta-train Optim4RL, $\text{LinearOptim}$, and $\text{L2LGD}^2$ in Humanoid and Ant; and Optim4RL is the only one that achieves satisfactory performance among them.

Next, we show that Optim4RL not only generalizes to unseen tasks, but also transfers to different hyper-parameter settings. 
To be specific, we train our learned optimizer Optim4RL under the default hyper-parameter setting and then test it under different hyper-parameter settings in two gridworlds -- \sds and \bdl. 
We report the returns at the end of training, averaged over 10 runs.
As shown in~\cref{tb:gae}, \cref{tb:entropy}, and~\cref{tb:discount}, Optim4RL is robust under different hyper-parameter settings, such as GAE $\lambda$, entropy weight, and discount factor.

\begin{table}[htbp]
\centering
\begin{tabular}{lcc}
\toprule
Task & Parameter Value & Return \\
\midrule 
\sds & 0.9   & 11.51$\pm$0.19 \\
\sds & 0.95  & 11.25$\pm$0.16 \\
\sds & 0.99  & 10.81$\pm$0.17 \\
\sds & 0.995 & 10.66$\pm$0.17 \\
\midrule
\bdl & 0.9   & 23.35$\pm$0.76 \\
\bdl & 0.95  & 23.57$\pm$0.60 \\
\bdl & 0.99  & 21.31$\pm$0.64 \\
\bdl & 0.995 & 20.55$\pm$0.54 \\
\bottomrule
\end{tabular}
\caption{The performance of Optim4RL with different \textit{GAE $\lambda$ values} in two gridworlds. All results are averaged over 10 runs, reported with $90\%$ confidence intervals.}
\label{tb:gae}
\end{table}

\begin{table}[htbp]
\centering
\begin{tabular}{lcc}
\toprule
Task & Parameter Value & Return \\
\midrule
\sds & 0.005 & 11.01$\pm$0.16 \\
\sds & 0.01  & 11.13$\pm$0.09 \\
\sds & 0.02  & 11.29$\pm$0.12 \\
\sds & 0.04  & 11.25$\pm$0.16 \\
\midrule
\bdl & 0.005 & 22.41$\pm$0.59 \\
\bdl & 0.01  & 22.45$\pm$0.79 \\
\bdl & 0.02  & 22.59$\pm$0.43 \\
\bdl & 0.04  & 19.96$\pm$1.30 \\
\bottomrule
\end{tabular}
\caption{The performance of Optim4RL with different \textit{entropy weights} in two gridworlds. All results are averaged over 10 runs, reported with $90\%$ confidence intervals.}
\label{tb:entropy}
\end{table}

\begin{table}[htbp]
\centering
\begin{tabular}{lcc}
\toprule
Task & Parameter Value & Return \\
\midrule 
\sds & 0.9   & 12.47$\pm$0.14 \\
\sds & 0.95  & 12.32$\pm$0.09 \\
\sds & 0.99  & 11.48$\pm$0.09 \\
\sds & 0.995 & 11.01$\pm$0.17 \\
\midrule
\bdl & 0.9   & 18.13$\pm$3.27 \\
\bdl & 0.95  & 25.01$\pm$1.42 \\
\bdl & 0.99  & 25.45$\pm$0.47 \\
\bdl & 0.995 & 22.07$\pm$0.81 \\
\bottomrule
\end{tabular}
\caption{The performance of Optim4RL with different \textit{discount factors} in two gridworlds. All results are averaged over 10 runs, reported with $90\%$ confidence intervals.}
\label{tb:discount}
\end{table}

\end{document}